\documentclass[sn-mathphys-ay]{sn-jnl}

\usepackage{graphicx}%
\usepackage{multirow}%
\usepackage{amsmath,amssymb,amsfonts}%
\usepackage{amsthm}%
\usepackage{mathrsfs}%
\usepackage[title]{appendix}%
\usepackage{xcolor}%
\usepackage{textcomp}%
\usepackage{manyfoot}%
\usepackage{booktabs}%
\usepackage{algorithm}%
\usepackage{algorithmicx}%
\usepackage{algpseudocode}%
\usepackage{listings}%
\usepackage{natbib}
\usepackage{comment}
\usepackage{wrapfig}
\usepackage{tabularx} 
\usepackage{tikz}
\usetikzlibrary{positioning}
\usepackage{enumitem}

\usepackage{subcaption}

\begin{document}

\title[Enhancing Interpretability in Generative Modeling: Statistically Disentangled Latent Spaces Guided by Generative Factors in Scientific Datasets]{\bf Enhancing Interpretability in Generative Modeling: Statistically Disentangled Latent Spaces Guided by Generative Factors in Scientific Datasets}

\author*[1]{\fnm{Arkaprabha} \sur{Ganguli}}\email{aganguli@anl.gov}

\author[2]{\fnm{Nesar} \sur{Ramachandra}}\email{nramachandra@anl.gov}

\author[3,4]{\fnm{Julie} \sur{Bessac}}\email{julie.bessac@nrel.gov}

\author[1]{\fnm{Emil} \sur{Constantinescu}}\email{emconsta@anl.gov}

\affil*[1]{\orgdiv{Mathematics \& Computer Science Division}, \orgname{Argonne National Laboratory}, \orgaddress{\city{Lemont}, \state{IL}, \postcode{60439}, \country{USA}}}

\affil[2]{\orgdiv{Computational Science Division}, \orgname{Argonne National Laboratory}, \orgaddress{\city{Lemont}, \state{IL}, \postcode{60439}, \country{USA}}}

\affil[3]{\orgdiv{Computational Science Center}, \orgname{National Renewable Energy Laboratory}, \orgaddress{\city{Golden}, \state{CO}, \postcode{80301}, \country{USA}}}

\affil[4]{\orgdiv{Department of Mathematics}, \orgname{Virginia Tech}, \orgaddress{\city{Blacksburg}, \state{VA}, \postcode{24061}, \country{USA}}}

\abstract{This study addresses the challenge of statistically extracting generative factors from complex, high-dimensional datasets in unsupervised or semi-supervised settings. We investigate encoder-decoder-based generative models for nonlinear dimensionality reduction, focusing on disentangling low-dimensional latent variables corresponding to independent physical factors. Introducing Aux-VAE, a novel architecture within the classical Variational Autoencoder framework, we achieve disentanglement with minimal modifications to the standard VAE loss function by leveraging prior statistical knowledge through auxiliary variables. These variables guide the shaping of the latent space by aligning latent factors with learned auxiliary variables. We validate the efficacy of Aux-VAE through comparative assessments on multiple datasets, including astronomical simulations.}




\maketitle

\section{Introduction}\label{sec:intro}

Semantic data representations are critical in artificial intelligence, significantly enhancing model performance in tasks like transfer and zero-shot learning \citep{lake2017building}. Central to this effort is to disentangle latent representations in generative models—representations where each latent dimension corresponds to an independent underlying factor of variation in the data. Disentanglement is achieved by leveraging statistical properties of the latent space and the dataset, enabling models where changes in one latent dimension affect only its corresponding factor without impacting others. This not only improves model interpretability but also enhances robustness against adversarial attacks \citep{Yang_Guo_Wang_Xu_2021}. For a comprehensive review of disentanglement and its statistical underpinnings, see \cite{disentangled-review}.

Datasets encountered in scientific research are often heterogeneous in modalities, fidelities, and accuracy where a particular entity or a state may be simultaneously associated with multiple images, graphs, vectors, scalar parameters, or labels with various associated measurement uncertainties. Besides, many natural and non-natural phenomena exhibit stochasticity, increasing the problem complexity. In many scientific problems, domain experts aim to understand and characterize underlying patterns and associations of physical quantities in order to improve their predictability for instance, or elucidate on the underlying physical phenomena. However, due to the problem complexity and data diversity (modality, fidelity, accuracy) these patterns are typically hard to extract from traditional data exploration tools. Moreover, domain experts are often cognizant of ``known knowns" and ``known unknowns", whereas several research problems also have associated ``unknown unknowns" \cite{Hatfield2022}. Classical data exploration rarely incorporates this type of partial or ``unknown unknowns" information, hence the need for novel tools as proposed here to advance science.  Finally, sensitivity analyses or computation of model response surfaces of input physical parameters are crucial for uncertainty quantification and forecasting \cite{razavi2016new, raghavan2018sensitivity}.

In Earth system science, for instance, information about a physical quantity can arise from numerical simulations, satellite imaging, or in situ sensors with various fidelities and uncertainties. For example, understanding the variation in weather patterns based on changing sea surface temperatures is essential for understanding the impacts of long-term environmental dynamics \cite{deser2014projecting, Maulik2020}. However, capturing the complex multi-scale variability of atmospheric phenomena remains an open challenge \cite{bauer2015quiet}, where data diversity along with partial expert knowledge is a typical setting, which remained un-leveraged by classical data science tools. Similar multi-modal, multi-fidelity datasets are also often encountered in astronomy: images and fluxes of galaxies are observed via telescopes, and a subset of the associated physical parameters, such as the stellar mass or galactic distances, are calculated \cite{bonvin2011galaxy}. Relationships between these factors provide valuable information about the evolution of galaxies over cosmic time \cite{newman2022photometric}, but they do not exhaustively explain all the physical processes and associations \cite{somerville2015physical}. In such studies, researchers might find it intriguing to use a generative model, where the latent factors are clearly disentangled with the `known knowns’ generative factors. Latent factors representing `known unknowns’ or `unknown unknowns’, however, might remain entangled, collectively contributing to the overall generation. Understanding these associations could aid domain experts in gaining deeper insights into the underlying mechanisms driving their data generation processes. In this paper, we demonstrate the applicability of our method on a representative galaxy catalog that encompasses both the data-level complexities and the desired science goals mentioned above.

Recent disentanglement research primarily explores unsupervised learning methods, introducing inductive biases into the Variational Autoencoder (VAE) framework to structure the latent space without using known factors of variation. Works such as \citep{betavae, beta-TCVAE, DIP_VAE} have advanced these techniques, which are detailed in Section \ref{sec:VAE_literature}. However, these methods often overlook auxiliary information that may be crucial in scientific datasets. Emerging semi/weakly supervised methods \citep{Chen_Batmanghelich_2020,  IDVAE} attempt to address this by leveraging observable ground-truth generative factors, but these approaches face challenges when auxiliary information is limited, requiring all ground-truth factors, which can be restrictive. To address this gap, we propose the \textit{\textbf{Aux}iliary information guided \textbf{V}ariational \textbf{A}uto\textbf{E}ncoder (\textbf{Aux-VAE})}, focusing on scenarios with available auxiliary information to disentangle representations with respect to known ground-truth generating factors while preserving data reconstruction capability. 


\textbf{Our contributions}: (i) \textbf{Statistically Interpretable Latent Space with Preserved Reconstruction Ability}: We partition the latent space into two segments to disentangle known factors of interest using auxiliary information. The first 
$d$ dimensions align with 
$d$ ground truth generative factors, enhancing interpretability, while the remaining dimensions capture other unknown factors in an entangled state to preserve overall accuracy. To achieve this, we construct a targeted prior and enforce disentanglement through posterior regularization. This method balances the trade-off between accurate data reconstruction and improved disentanglement.
(ii) \textbf{Enabling Control Over Ground Truth Factors for Understanding Dataset Characteristics}: Our disentanglement approach establishes a direct correspondence between the disentangled latent factors and the known generative factors of interest via the specification of the latent factor's distributions. This facilitates a precise understanding of specific physical characteristics of the data by reconstructing it with the corresponding latent factors adjusted to reflect the controlled level of the generative factors. This also enables a computationally efficient sensitivity study where one can compute model responses across the space formed by ground truth factors. (iii) \textbf{Introducing a Novel Disentanglement Metric}: We introduce a new metric that provides an intuitive, cost-effective way to measure disentanglement in the latent space relying on correlations between latent factors. This metric avoids the need for retraining separate models and quantitatively assesses the qualitative aspects of disentanglement, scoring up to 1 for optimal separation of factors.
The article outlines our methods and results as follows. Section \ref{sec:VAE_literature} formalizes key definitions and reviews relevant literature; Section \ref{sec:proposed_method} introduces the Aux-VAE methodology; Section \ref{sec:experiments} presents experiments on both scientific and benchmark datasets; and Section \ref{sec:conclusion} offers conclusions and future directions.

\section{Variational Autoencoders (VAE) and related literature}\label{sec:VAE_literature}
We begin with a generative model for observed data, where a latent variable $z$ is sampled from $p(z)$, and an observation $x$ is generated by sampling from $p_{\theta}(x|z)$. The joint density of the latent variables and observations is denoted as $p_{\theta}(x, z) = p(z)p_{\theta}(x|z)$. The inference problem involves computing the posterior of the latent variables conditioned on the observations, i.e., $p_{\theta}(z|x) = \frac{p_{\theta}(x,z)}{ \int p_{\theta}(x,z) dz}$. Given a finite set of samples (observations) from the true data distribution $p(x)$, the exact computation of the posterior is generally intractable and requires approximate inference. Variational inference addresses this by positing a family of approximate densities $\mathcal{Q}$ over the latent factors and minimizing the Kullback-Leibler (KL) divergence to the true posterior, i.e., $q^*_x = \min_{q \in \mathcal{Q}} \mathrm{KL}(q(z)||p_{\theta}(z|x)$) \citep{blei2017variational}. A variational autoencoder (VAE) utilizes the amortized inference,  a recognition model, parameterized by $\phi $ to encode an inverse map from observations to approximate posteriors. The recognition model parameters are learned by optimizing the problem $\min_{\phi} E_x[\mathrm{KL}(q_{\phi}(z|x)||p_{\theta}(z|x))]$, where the outer expectation is over the true data distribution $p(x)$ from which we have samples. This optimization is equivalent to maximizing the Evidence Lower Bound (ELBO):
\begin{equation}\label{eq:VAE_ELBO}
\arg \min_{\phi} \mathbb{E}_x[\mathrm{KL}(q_{\phi}(z|x)||p_{\theta}(z|x))] = \arg \max_{\phi} \mathbb{E}_x \mathbb{E}_{z \sim q_{\phi}(z|x)} [\log p_{\theta}(x|z)] - \mathrm{KL}(q_{\phi}(z|x)||p(z))
\end{equation}
Often, the density forms of $p(z)$ and $q_\phi(z|x)$ are chosen such that their $\mathrm{KL}$-divergence can be written in a closed-form expression (e.g.,
$p(z)$ is $\mathcal{N}(0, I)$ and $q_\phi(z|x)$ is $\mathcal{N}(\mu_\phi(x), \Sigma_\phi(x))$) \citep{kingma2022autoencoding}. This framework encourages the encoder to learn meaningful representations in the latent space while enabling the decoder to generate data samples that closely match the input data.

\textbf{Unsupervised approaches}: In the context of representation learning, all the unsupervised approaches follow the basic structure of the VAE and enforce the desired disentangling characteristic in the latent space with an additional term in the loss function. For example, $\beta-$VAE \cite{betavae} imposes a weight  $\beta > 0$ to the $\mathrm{KL}$-term in the VAE-objective to ensure that each latent factor captures independent sources of variation in a disentangled manner.
Building upon this, \cite{factor_VAE} introduce an additional term in the VAE loss function, $\mathrm{KL}(q(\mathbf{z})||\prod_{j}q(z_j))$, encouraging the aggregate posterior $q(\mathbf{z})$ to align with the product of marginals $q(z_j)$, thereby achieving independent latent factors, originally referred to as Total Correlation (TC) in the literature. Numerous representation learning algorithms \cite{DIP_VAE, factor_VAE} have been proposed in this unsupervised fashion, proving beneficial in scenarios where auxiliary information on the ground truth factors is unavailable. However, a plausible drawback of unsupervised methods is the identifiability issue, where different models may yield entirely different latent variables despite having the same marginal data and prior distributions due to potential transformations of the latent variable while preserving the marginal distribution. To address this, posterior regularization via the choice of variational family and the prior distribution has been investigated in the literature \cite{mathieu2019disentangling, kumar2020implicit}. Nonetheless, a common limitation of unsupervised approaches is their tendency to exhibit high variance, making it challenging to identify well-disentangled models without supervision \citep{locatello2019fairness}. This is consistent with the theoretical findings of \cite{locatello2019challenging}, suggesting that unsupervised learning of disentangled representations is unfeasible without appropriate inductive biases.

\textbf{Disentanglement with auxiliary information}: In response to certain limitations, a category of methods has emerged that leverages auxiliary information about ground truth factors within the traditional VAE framework. For example, existing semi-supervised approaches tackle the disentanglement of observed factors by utilizing limited supervised data on class levels (\cite{Cars3d-paper, cheung2014discovering, mathieu2016_neurips, paige2017learning, kingma2014semi}). State-of-the-art weakly-supervised disentanglement methods operate under the assumption that observations are grouped based on known relationships between images within the same group and their corresponding groups (\citep{bouchacourt2018multi, hosoya2018group, Chen_Batmanghelich_2020, locatello2020weakly}). A concise overview of these methodologies can be found in \cite{shu2019weakly}. While these approaches have demonstrated success in computer vision and other scientific domains, they face challenges in scenarios where generating factors are continuous and multiple sources of true generative factors remain unknown. This property is often seen in a wide range of scientific datasets, from astrophysics (both in observational datasets such as the COSMOS \cite{cosmos2007} or simulated datasets created for telescope surveys \cite{korytov2019cosmodc2}) to earth system studies (e.g., \cite{kaltenborn2023climateset}) and medical sciences (e.g., \cite{efron2004least}). 
In the context of such datasets, clustering the images into distinct groups (or any other operation on the latent space of the VAE -- such as classification, regression, or anomaly detection) is difficult if auxiliary information is not used along with a robust disentanglement scheme. 

Taking a step towards a more general setting, \cite{khemakhem2020variational} introduced the Identifiable-VAE (IVAE) framework. This framework learns a disentangled representation by employing a factorized prior from the exponential family, conditioned on auxiliary variables representing certain generative factors. Building upon this foundation, \cite{IDVAE} proposed an iterative training strategy `IDVAE' utilizing two VAEs: one to capture latent representations from auxiliary information and the other to leverage these latent distributions for learning the data distribution. Despite the appealing theoretical guarantee of identifiability, \cite{kim2023covariate} observed that IVAEs may overlook observations in certain cases, potentially leading to posterior collapse in experiments. As discussed in \cite{DIP_VAE}, one potential remedy for this issue involves imposing regularization on the Expected variational posterior. We extend this approach by incorporating limited auxiliary information on the ground-truth factors, as elaborated in Section \ref{sec:proposed_method}.

\section{Limited available auxiliary information and proposed approach}\label{sec:proposed_method}
\textbf{Observed database}: Moving forward, let us presume access solely to a subset of the ground truth factors  $S_{obs}\in \mathbb{R}^d$, conveyed through auxiliary variables $u \in \mathbb{R}^d$. Each auxiliary variable is intended to encapsulate a specific ground truth factor within $S_{obs}$. Our aim lies in disentangling the latent space concerning these identified ground truth factors $S_{obs}$, observable via $u$. We initiate this endeavor with an observed database $\mathcal{D}$  comprising $n$ independent and identically distributed pairs of $x$ and $u$, denoted as $\mathcal{D} = \{(x^{(1)}, u^{(1)}), (x^{(2)}, u^{(2)}), \dots, (x^{(n)}, u^{(n)})\}$. 
We begin by reformulating the ELBO of a VAE:
\begin{align}\label{eq:ELBO}
L_{VAE}= \mathbb{E}_{z \sim q(z|x)}\left[\log p(x|z)\right] - \mathrm{KL}\left(q(\mathbf{z}|\mathbf{x}) | |p(\mathbf{z})\right),
\end{align}
where, $q(z|x)$ denotes the encoder, $p(x|z)$ represents the decoder, and $p(z)$ is the prior distribution. To incorporate information from auxiliary variables, we partition the latent space into two distinct components: auxiliary-informed latent factors $z_{aux}$ and residual latent factors $z_{recon}$, represented as: $ z^{1 \times d_Z}=\left( z_{aux}^{1 \times d}, z_{recon}^{1 \times (d_Z-d)}\right)$.
Here, the auxiliary-informed latent factors $Z_{aux}$ signify the latent features associated with auxiliary variables in a disentangled fashion, while the residual latent factors $Z_{recon}$ characterize the latent features necessary to capture the underlying factors not explicitly covered by the auxiliary variables $u$.  Hence, conditional on the auxiliary variables $u$, we define the prior in the following way: 
\begin{align}\label{eq:prior}
    & p_{z|u}(z)=  \left(\prod_{j=1}^{d} p_{\mathcal{N}(u_j,\frac{1}{n})}(z_j)\right) p_{\mathcal{N}(0, I_{d_Z - d})}(z_{(d+1):d_Z}) = p_{\mathcal{N}(\mu_0,\Sigma_0)} (z)
\end{align}

where $\mu_0= (u_1, u_2, \dots, u_d, 0, \dots, 0)$ and $\Sigma_0=diag(\frac{1}{n} I_{d}, I_{d_Z - d})$ denotes the mean and variance of the Gaussian distribution in the prior.  Now, while optimizing the ELBO in Eq. \eqref{eq:ELBO} (derivation detailed in the SM), the first term can be estimated by simple Monte-Carlo approximation: $\mathbb{E}_{z \sim q(z|x)}\left[\log p(x|z)\right] \hat{=} \frac{1}{J}\sum\limits_{j=1}^J log(p(x|z^j))$. 

Now, the $\mathrm{KL}$ part can be decomposed as follows utilizing the closed-form structure of the Gaussian distributions: 
\begin{align}
    \mathrm{KL}\left(q(\mathbf{z}|\mathbf{x}) | |p(\mathbf{z|u})\right) & =   \mathrm{KL}\left(\mathcal{N}(\mu_\phi, \Sigma_\phi) || \mathcal{N}(\mu_0, \Sigma_0)\right) \nonumber\\
    & = \frac{1}{2}\left[log\frac{|\Sigma_0|}{|\Sigma_\phi |} -d_Z +(\mu_\phi-\mu_o)'\Sigma_0^{-1}(\mu_\phi-\mu_o) + tr(\Sigma_0^{-1} \Sigma_\phi)\right]
    .
\end{align}

\paragraph{Advancing Disentanglement via Expected Variational Posterior:} Achieving disentangled latent spaces requires more than aligning the encoder distribution $q_\phi(z|x)$ with the desired prior. Disentanglement should also be fostered in the \textit{expected posterior} $p_\theta(z) = \int p_\theta(z|x)p(x)dx$. Its variational counterpart is expressed as the inferred prior or expected variational posterior: $q_\phi(z)=\int q_\phi(z|x)p(x)dx$. 
 Utilizing the pairwise convexity property of $\mathrm{KL}$-divergence, we can show that the distance between $q_\phi(z)$ and $p_\theta(z)$ is bounded by the objective of the variational inference \cite{DIP_VAE}:
\begin{align}\label{eq:KL_variational_posterior}
    \mathrm{KL}(q_\phi(z)||p_\theta(z)) & = \mathrm{KL}\left(E_{x \sim p(x)}q_\phi(z|x)||E_{x \sim p(x)}p_\theta(z|x)\right) \nonumber\\
    & \leq E_{x \sim p(x)} \mathrm{KL}(q_\phi(z|x)||p_\theta(z|x)).
\end{align}
Hence, although, maximizing ELBO \eqref{eq:ELBO} would ideally decrease $\mathrm{KL}(q_\phi(z)||p_\theta(z))$, in many complex scenarios, the two sides of equation \eqref{eq:KL_variational_posterior} might deviate at the stationary point of convergence \citep{DIP_VAE}. Explicitly minimizing $\mathrm{KL}(q_\phi(z)||p(z))$ provides better control over the disentanglement. However, due to the intractable $\mathrm{KL}$ term, we implicitly enforce the following three main characteristics of the disentangled prior concerning auxiliary information $u$, such as: (1) \textbf{\textit{Inter-independence}: }$u \perp z_{recon}$, (2) \textbf{\textit{Intra-independence}:} $u_{j} \perp z_{aux,j'}$ for $j,j'=1,2,\dots d, j \neq j'$, and (3) \textbf{\textit{Explicitness}:} $E_{z \sim p(z|u)} (z_{aux,j})=u_j$ for $j=1,2,\dots d$. 

Quantifying the interdependency among $z_{aux}$, $z_{recon}$, and the auxiliary information $u$ presents a challenge. To address this, we turn to polynomial regression, a technique that assesses nonlinear relationships among variables \cite{poly_reg}. This approach measures the strength of dependency by aggregating correlations across various polynomial degrees and utilizes Monte Carlo samples of encoder outputs to compute these correlations.
Specifically, to calculate these correlations, we utilize the expected latent factors $\mu_\phi$ from the encoder $q_\phi(z|x)$. This choice is justified by the theorem of total variance. For instance, in the case of \textit{Inter-independence}, the covariance between the polynomials $u^k$ and $z_{recon}^{k'}$ can be expressed as:
\begin{align}\label{eq:total_variance}
Cov(u^k, z_{recon}^{k'}) & =  E_{(x,u) \sim p(x,u)}Cov_{u, z \sim q_\phi (z|x)}(u^k, z_{recon}^{k'}) \nonumber\\ 
& \hspace{1cm} + Cov_{(x,u)} (u^k, E_{z \sim q_\phi(z|x)}(z_{recon}^{k'}) \nonumber\\
& = Cov_{(x,u)} (u^k, \mu_{\phi, d+1:d_Z}^{k'}).
\end{align}
In this expression, under the outer expectation, the covariance in the first term becomes zero conditioned on $u$. Analogous properties apply to \textit{Intra-independence} and \textit{Explicitness} as well (detailed in Section 1 of the SM). For notational simplicity, let us denote, $\mu_{\phi, 1:d}=\mu_{\phi, aux}$ and $\mu_{\phi, d+1:d_Z}=\mu_{\phi, recon}$.

With this approach, for two random vectors $v^{m_v \times 1}$ and $w^{m_w \times 1}, m_v \leq m_w$, we define the correlation matrix $Corr(v,w)$ as $diag(\Sigma_{v,v})^{-1/2}\Sigma_{v,w} diag(\Sigma_{w,w})^{-1/2}$, with $\Sigma_{v,w}^{m_v \times m_w}=\mathbb{E}[(v-\mathbb{E}(v))(w-\mathbb{E}(w))']$. Subsequently, we formulate the following two dependency metrics: 

\begin{align}
& R^K_0(v,w)= \frac{1}{K m_v m_w}\sum\limits_{k,k'=1, k\neq k'}^K \sum\limits_{i=1}^{m_v}\sum\limits_{j=1}^{m_w}|\left(Corr(v^k, w^{k'})\right)_{ij}|, \\
& R^K_1(v,w)= \frac{1}{K m_v m_w}\sum\limits_{k,k'=1, k\neq k'}^K \sum\limits_{i=1}^{m_v}\left(1-|\left(Corr(v^k, w^{k'})\right)_{ii}| \right).
\end{align}

In these metrics, the first summation aggregates associations from all possible polynomial combinations up to degree $K$, while the second sum separately considers various terms of the covariance matrix in $R_0(\cdot, \cdot)$ and $R_1(\cdot, \cdot)$. Consequently, $R_0(\cdot, \cdot)$ and $R_1(\cdot, \cdot)$ quantify the strength of pairwise nonlinear dependency by evaluating the association among the polynomials of the variables, enhancing our understanding of disentanglement.

These metrics are then incorporated into the loss function to enforce optimal disentanglement within and between $Z_{aux}$ and $Z_{recon}$. Specifically, three regularization terms are introduced into the objective function:
\begin{align}\label{eq:main_objective}
    \mathcal{L}_{Aux-VAE}=\mathcal{L}_{VAE} &+ \underbrace{\lambda_1\sum\limits_{j=1}^d \Big(R^K_1(u_j,\mu_{\phi, aux,j}) + R^K_0(u_j,\mu_{\phi, aux,-j})\Big)}_{\text{Intra-independence and explicitness regularizer}} \nonumber \nonumber \\ 
    &+ \underbrace{\lambda_2\Big(R^K_0(u, \mu_{\phi, rec})\Big)}_{\text{Inter-independence regularizer}} 
\end{align}
Here, the three regularizers play a crucial role and align with the intuitive logic of achieving disentanglement. The intra-group regularization includes two terms: the first ensures that each dimension of $z_{aux}$ closely aligns with the auxiliary information $u$, while the second imposes a penalty on the dependency between any two latent factors in $z_{aux}$ using the defined polynomial dependency metric. Similarly, the inter-group regularization aims to reduce the dependency between $z_{aux}$ and $z_{recon}$. No restrictions are imposed on the dependency within $z_{recon}$ to ensure good reconstruction quality.

\section{Experiments}\label{sec:experiments}
\subsection{Experimental settings}\label{sec:exp_setting}
We compare the proposed approach against two major alternative disentanglement methods: $\beta$-VAE \citep{betavae}, and IDVAE \citep{IDVAE}.
These two methods were chosen to represent different classes of existing disentangling approaches.
$\beta$-VAE \citep{betavae} serves as a baseline for its simple yet effective unsupervised approach with minimal assumptions. It represents the class of unsupervised methods utilizing no ground-truth factor for disentanglement, but a regularization term is introduced to enforce disentanglement.
On the other hand, IDVAE \citep{IDVAE} represents recent algorithms on auxiliary variable-informed methods, which is the most closely related approach in the literature analogous to Aux-VAE.
We implemented these methods using the same hyperparameter settings as their publicly available repositories. All methods were implemented in PyTorch \citep{pytorch}, with code available at \cite{AuxVAE2024}.


\subsubsection{Datasets} We have created a representative scientific dataset where measurements from instruments 
such as telescopes are associated with physical quantities. We simulate the galaxy images observed from telescopes using GalSim \cite{Galsim2015}, a widely used in current and future space- and ground-based telescope missions \cite{LSST2023, DES2022, Euclid2023}. Each image is associated with 5 physical parameters, Apparent brightness of the galaxy ($flux$, in number counts), radius of the galaxy ($radius$, in arc-seconds), 2 reduced gravitational shear components ($g1$ and $g2$ in Cartesian coordinates), and the full width of half maximum of the Gaussian function (also called point-spread function, $psf$) used in the convolution. Further details of the experimental design are provided in the SM Section 1.

To maintain experiment realism, we refrain from using all ground-truth information as auxiliary variables. Instead, we compare results under a more
\begin{wrapfigure}{r}{0.8\textwidth}
      \centering
     \includegraphics[height=3cm,width=10cm]{  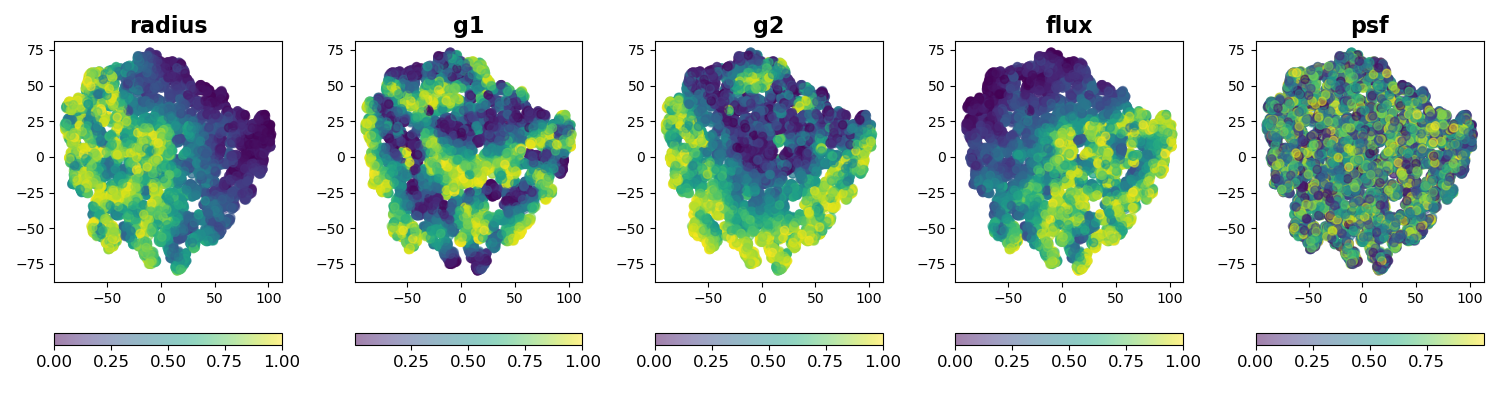}
      \caption{t-SNE plot of the galaxy images colored by each of the generating factors, highlighting $psf$ as the least important generator in this dataset.}
      \vspace{-3pt}
      \label{fig:t-sne-x-vs-u}
\end{wrapfigure}
 practical setting. Specifically, we categorize the auxiliary variables into important and less important categories. The quantities $radius$, $g1$, and $g2$ as well as $flux$ are considered important, as they focus on essential physical characteristics, while $psf$ is deemed less significant. This hierarchy of importance is illustrated in Figure \ref{fig:t-sne-x-vs-u}. While this heuristic characterization is based on domain expertise, one may investigate correlation structures to identify auxiliary variables.  
In light of this, we examine the following three cases: \textbf{Case 1:} Using all five ground truth generating factors as auxiliary information, \textbf{Case 2:} Using only the important factors ($radius$, $g1$, and $g2$) as auxiliary information, excluding $flux$ and $psf$, and \textbf{Case 3:} Using mostly the less important factors ($flux$, and $psf$) as auxiliary information, omitting $radius$, $g1$, and $g2$.
Evaluating these cases illustrates the interplay between the $Z_{aux}$ and $Z_{recon}$ factors, offering insights for datasets where exhaustive ground-truth factors are unknown. All ground-truth factors were normalized within the range [0, 1], assuming an implicit ordering for discrete factors before normalization.

Furthermore, to showcase the consistency and efficacy of our proposed method using auxiliary information, we conduct experiments on two other synthetic datasets: `Cars3D' \citep{Cars3d-paper} and `DSprites' \citep{beta-vae-main}. These datasets provide explicit access to ground-truth factors, allowing us to assess our method against competing approaches. 

\subsubsection{Disentanglement metric} Assessing the degree of disentanglement among latent factors is crucial in understanding the effectiveness of a generative model, yet a nontrivial task in practice. While various disentanglement metrics exist in the literature \cite{betavae, factor_VAE, DIP_VAE}, many rely on fitting a supervised regression between the learned latent space and ground-truth factors. However, implementing additional regression models for evaluation can incur significant computational costs. Moreover, studies have shown that these model-based metrics may not always correlate well with qualitative disentanglement observed in latent traversal plots \citep{DIP_VAE}. To address these challenges and efficiently evaluate disentanglement, we propose a novel metric called the Linear Disentanglement Score (LDS). For the auxiliary features $u_j$'s and the latent factors $z_l$'s, the LDS is defined as: 

\begin{equation}\label{eq:LDS}
    LDS = \frac{1}{d}\sum\limits_{j=1}^{d}\frac{\max\limits_l Corr(u_j, z_l)}{\sum\limits_{l=1}^{d_Z}|Corr(u_j, z_l)|}.
\end{equation}

The underlying idea stems from the concept that each generative factor $u_j$ should ideally correlate with only one latent factor $z_l$. Therefore, for any $u_j$, a value closer to 1 for the term inside the summation indicates better disentanglement among the latent factors. Through a straightforward mathematical rationale, we establish that $LDS \in [\frac{1}{d_Z},1]$. While the idea of measuring the strength of linear dependence for this purpose is not new. The SAP score \citep{DIP_VAE} is computed as the average difference in prediction accuracy between the most and second-most predictive latent dimensions for each generative factor.  A higher SAP indicates that a factor is captured predominantly by a single latent variable, reflecting strong disentanglement.  For continuous factors, SAP ranges in \((0,1]\).  For categorical factors, balanced classification accuracy is used, allowing SAP to exceed 1. Our LDS metric offers an extension, providing a bounded version of the SAP-score for evaluating disentanglement. Moreover, it is crucial to recognize that a high SAP score might not exclude the possibility of one latent dimension effectively capturing multiple generative factors. Conversely, our proposed metric emphasizes evaluating each latent factor's capacity to represent individual generative factors distinctly or become entirely independent, thus offering a more comprehensive assessment of disentanglement.

\subsection{Experimental results}\label{sec:exp_results}
In this section, we present the results of our numerical experiments, focusing on four main criteria: (1) Reconstruction accuracy, (2) The relative importance of $Z_{aux}$ and $Z_{recon}$ in the overall reconstruction,  (3) Disentanglement among the latent factors, (4) Latent traversal (generating from the learned decoder network by varying only one latent while keeping the others fixed)  across Cases 1, 2, and 3 for the galaxy dataset mentioned in Section \ref{sec:exp_setting}. To highlight the versatility of our method, we present results from Aux-VAE on both a galaxy simulation dataset and non-scientific datasets like Cars3D' and DSprites'. Primarily, we focus on the galaxy dataset due to space constraints, with further results available in the supplementary materials.  

\subsubsection{Reconstruction accuracy}
It is common to encounter a trade-off between reducing reconstruction error and
  \begin{wrapfigure}{r}{0.6\textwidth}
      \centering
     \includegraphics[height=5cm,width=6cm]{  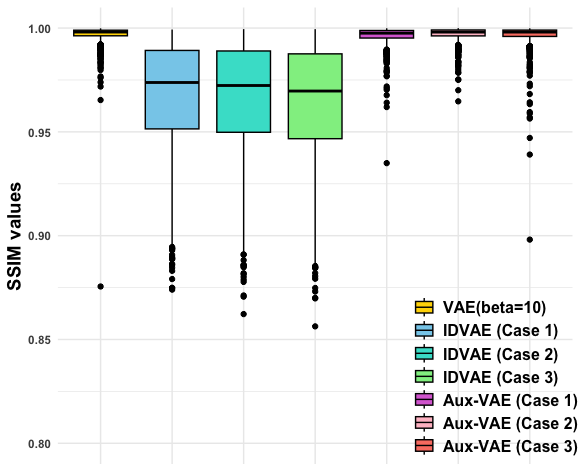}
      \caption{Distribution of SSIM between the original and reconstructed test images.}
      \vspace{-20pt}
      \label{fig:reconstruction_SSIM_all}
\end{wrapfigure} inducing disentanglement through posterior regularization. As a result, achieving disentanglement often comes at the cost of increased reconstruction error. Therefore, we use $\beta-$VAE as a baseline for assessing reconstruction accuracy. 
For evaluation, we employ the structural similarity index measure (SSIM) \citep{SSIM-paper, SSIM-python}, a metric adept at quantifying image similarity by considering luminance, contrast, and structure.   The SSIM between two image patches \(x\) and \(y\) is defined as
  \[
    \mathrm{SSIM}(x,y) \;=\; 
      \frac{\bigl(2\mu_x\mu_y + C_1\bigr)\,\bigl(2\sigma_{xy} + C_2\bigr)}
           {\bigl(\mu_x^2 + \mu_y^2 + C_1\bigr)\,\bigl(\sigma_x^2 + \sigma_y^2 + C_2\bigr)},
  \]
  where \(\mu_x, \mu_y\)\ are the local means of \(x\) and \(y\), \(\sigma_x^2, \sigma_y^2\)\ are the local variances, \(\sigma_{xy}\)\ is the local covariance between \(x\) and \(y\), \(L\)\ denotes the dynamic range of the pixel values (e.g.\ 255 for 8-bit images), \(C_1 = (K_1 L)^2\) and \(C_2 = (K_2 L)^2\)\ are stabilizing constants (typically \(K_1 = 0.01\) and \(K_2 = 0.03\)). In our experiments, SSIM is computed via the Python library \texttt {skimage.metrics.structural\_similarity} \citep{scikit-image}.
With SSIM scores ranging between -1 and 1, where 1 signifies perfect similarity, 0 denotes no similarity, and -1 implies perfect anti-correlation, it offers a nuanced assessment compared to pixel-wise methods. Figure \ref{fig:reconstruction_SSIM_all} presents the SSIM scores for all methods across Cases 1, 2, and 3 using the galaxy image dataset.

Aux-VAE maintains strong reconstruction capabilities, on par with the baseline VAE, especially in Case 3, where key factors are missing from the auxiliary information, by effectively leveraging the residual latent factors $Z_{recon}$. Although IDVAE displays similar overall performance, it experiences a noticeable reduction in accuracy from Case 1 to Case 3, due to the exclusion of many crucial generative factors from the auxiliary information and the absence of other latents in their implementation to compensate for missing information.

\subsubsection{Disentanglement among the latent factors with respect to the ground-truth generating factors}
Building on the fundamental concept of `disentanglement' outlined in Section \ref{sec:intro}, where each latent factor is expected to correspond to a single underlying generative factor, we approach this comparison from two distinct angles. \textbf{Quantitatively}, we calculate the LDS metric \eqref{eq:LDS} and SAP score \cite{DIP_VAE} across all cases and datasets, as summarized in Table \ref{tab:LDS_all}. Aux-VAE achieves notably high LDS scores, reflecting its strong ability to capture underlying relationships with robust disentanglement, as further supported by the reconstruction quality in Figure \ref{fig:reconstruction_SSIM_all}. While the SAP score shows a similar trend, it tends to be higher when multiple latent factors are related to the same generating factor, a known limitation \citep{DIP_VAE}.
\begin{table}[!h]
    \centering
    \caption{Disentanglement Comparison Across Methods on Various Datasets and Cases Using the Proposed LDS Metric (SAP Scores in Parentheses).}
    \label{tab:LDS_all}
        \begin{tabular}{|c|c|c|c|c|}
            \hline
            \textbf{Dataset} &  & \textbf{$\beta$-VAE} & \textbf{IDVAE} & \textbf{Aux-VAE} \\ \hline
            \multirow{3}{*}{Galaxy image} & Case 1 & 0.48 (0.39) & 0.65 (0.72) & \textbf{0.88 (0.81)} \\ 
             & Case 2 & - & 0.73 (0.74) & \textbf{0.94 (0.89)} \\ 
             & Case 3 & - & 0.59 (0.68) & \textbf{0.81 (0.84)}\\ \hline
            Card3D & - & 0.21 (0.18) & 0.67 (0.55) & \textbf{0.93 (0.85)}\\ \hline
            DSprites & - & 0.26 (0.27) & 0.39 (0.43) & \textbf{0.83 (0.78)}\\ \hline
        \end{tabular}
\end{table}

\begin{figure}[!t]
      \centering
     \includegraphics[height=4.5cm,width=13cm]{  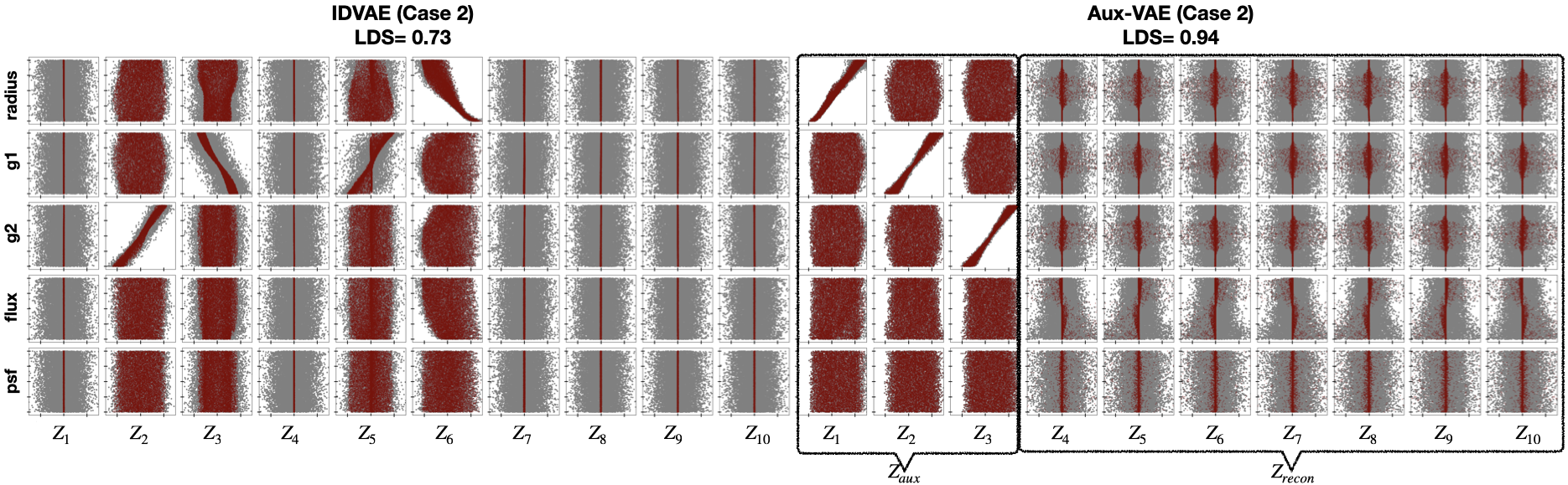}
      \vspace*{-5mm}
      \caption{Visualizing Disentanglement: Latent factors are displayed along the x-axis and generative factors along the y-axis. This scatterplot contrasts latent factors ($Z$, represented by grey dots) with the latent means ($\mu_\phi$, shown as maroon dots), alongside highlighting the LDS metric. }
      \label{fig:Z-vs-U_all}
\end{figure}

From a more \textbf{qualitative} standpoint, we examine each latent factor with respect to the underlying generative factors. Figure \ref{fig:Z-vs-U_all} showcases Case 2 for the galaxy image dataset, with similar plots for other datasets provided in the SM. We observe that Aux-VAE's auxiliary-informed latent factors $Z_{aux}$ exhibit clear associations with the corresponding generating factors, remaining independent of the remaining latent factors. In Case 2, the residual latent factors $Z_{recon}$ are less significant, since the auxiliary information encompasses most of the crucial ground-truth generating factors. Nonetheless, it is apparent that they are collectively attempting to represent the $flux$ in an entangled manner. 
Meanwhile, IDVAE also demonstrates some degree of one-to-one association with the associated factors. However, these relationships are often entangled, making it challenging to definitively attribute one particular latent factor to a specific generating factor. Despite IDVAE's factorized prior promoting independence in the latent space, the collective impact of entire auxiliary information on each dimension of the latent distribution may hinder complete disentanglement, especially for interdependent generative factors. A similar pattern is observed for basic $\beta-$VAE (relegated to the SM), where representations appear to be even more entangled.

\subsubsection{Relative importance between $Z_{aux}$ and $Z_{recon}$}
The proposed method, Aux-VAE, relies heavily on the interplay between two classes of latent factors, $Z_{aux}$ and $Z_{recon}$. While the former aims to capture limited information 
\begin{wrapfigure}{r}{0.5\textwidth}
      \centering
     \includegraphics[height=6cm,width=5cm]{ 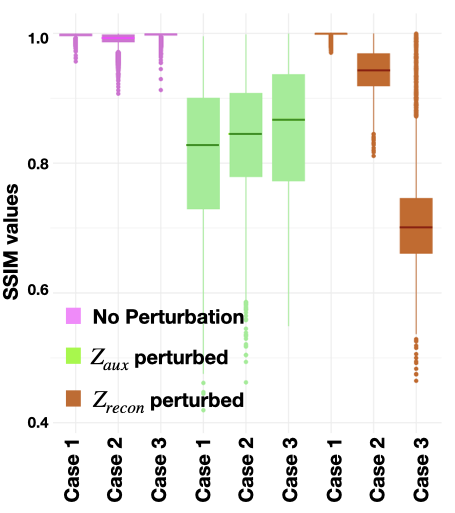}
      \caption{Assessing Latent Factor Importance: $Z_{aux}$ vs $Z_{recon}$ Analysis with Gaussian Noise Perturbation.}
      \vspace{-10pt}
      \label{fig:Z-aux-vs-Z_recon_imp}
\end{wrapfigure}
from the auxiliary data in a disentangled manner, the latter endeavors to 
reconstruct the remaining unknown generative factors, not covered by the auxiliary data, in an entangled manner to achieve better reconstruction. Therefore, in scenarios where
 the majority of generative factors remain unknown and are absent from the auxiliary data, $Z_{recon}$ becomes increasingly important in understanding the overall data generation process.

To empirically demonstrate this, we consider three cases for the galaxy images dataset outlined in Section \ref{sec:exp_setting}. To assess the relative importance of the latent factors, we conduct the following experiment: 
For each case, we utilize the encoder model on 1000 test images to obtain their latent representations $Z^{test}=(Z_{aux}^{test}, Z_{recon}^{test})$ and use the decoder to generate the outputs. Then, to understand the importance of $Z_{aux}$, we perturb only the factors in $Z_{aux}^{test}$ with additive Gaussian noise, creating $Z_{aux}^{perturbed}$, and reconstruct the images using the decoder with $Z^{perturbed}=(Z_{aux}^{perturbed}, Z_{recon}^{test})$.
Similarly, to assess the importance of $Z_{recon}$, we conduct a similar experiment but perturb $Z_{recon}^{test}$ instead. The SSIM boxplots for the 1000 test images are illustrated in Figure \ref{fig:Z-aux-vs-Z_recon_imp}. We denote the first reconstruction with $Z^{test}=(Z_{aux}^{test}, Z_{recon}^{test})$  as `No perturbation,' and the subsequent experiments as `$Z_{aux}$ perturbed' and `$Z_{recon}$ perturbed.' As expected, under the `No perturbation' (magenta) setting, all three models perform similarly. However, with perturbations, a decrease in reconstruction quality due to the disruption of a significant latent factor will result in lower SSIM values, signaling the greater importance of that factor.
In the `$Z_{aux}$ perturbed' (green) setting, we observe a gradual increase in SSIM from Case 1 to Case 3. In Case 1, where all true generating factors are included in the auxiliary information, $Z_{aux}$ is most sensitive to perturbations compared to Cases 2 and 3. Conversely, under `$Z_{recon}$ perturbed' (maroon) setting, we observe no sensitivity in Case 1, while in Case 3, where most important generative factors are unavailable in the auxiliary data, $Z_{recon}$ shows high sensitivity, resulting in a significant drop in SSIM. This experiment validates the intuition behind Aux-VAE's latent structure formation and underscores the importance of $Z_{recon}$ in more realistic scenarios where most of the true generative factors are unknown.

\subsubsection{Latent space traversal - systematically changing each dimension of $Z_{aux}$ at a time}
\begin{figure}[t]
      \centering
     \includegraphics[height=5.5cm,width=13cm]{ 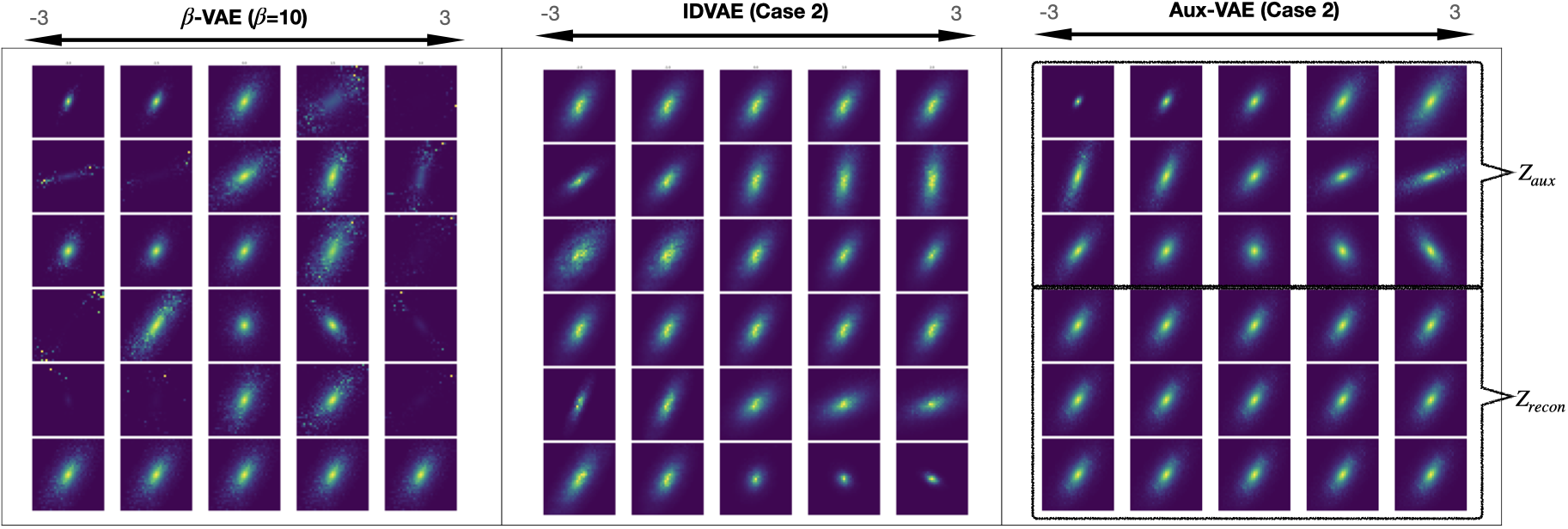}
      \vspace*{-4mm}
      \caption{Latent space traversal - Visualization of Latent Factor Adjustments in VAE, IDVAE, and Aux-VAE, showing how systematic changes to individual factors impact generated galaxy images.}
      \label{fig:recon_from_latent}
\end{figure}

In our exploration of the latent space traversal, we aim to unravel the influence of individual latent factors on the generated outputs. Beginning with the extraction of latent factors from a sample image using the encoder network, we initiate the traversal by incrementally adjusting one latent factor's values at a time while keeping the others constant. Leveraging the decoder network, we then produce the corresponding output for each adjustment, visualizing the results along a row in Figure \ref{fig:recon_from_latent}. This systematic exploration allows us to gain insight into how variations in latent factors affect the generated images. 
Figure \ref{fig:recon_from_latent} features three models—$\beta$-VAE, IDVAE, and our proposed Aux-VAE, used in case 2 of the galaxy image dataset. Due to space limitations, we present the latent space traversal for only the top six most sensitive latent factors out of ten for each method. Detailed traversal results for this and additional datasets are provided in the SM. 

We note that Aux-VAE’s auxiliary latent factors $Z_{aux}$ effectively adapt to the corresponding generating factors like $radius$, $g1$, and $g2$, thereby maintaining the geometric significance of these auxiliary factors. Given the minimal importance of $Z_{recon}$ in this scenario, no noticeable changes are observed in the generated images during latent space traversal for $Z_{recon}$. This underscores the model's capacity to encapsulate relevant auxiliary information in a disentangled fashion.
In contrast, for VAE and IDVAE, the latent factors exhibit more entangled relationships, as observed in the scatterplot shown in Figure \ref{fig:Z-vs-U_all}. However, in the latent space traversal experiment, IDVAE performs better than VAE, likely due to the encompassing of auxiliary information. Specifically, the fifth, second, and eighth latent factors in IDVAE align well with the underlying generative factors $radius$, $g1$, and $g2$, respectively.
\section{Conclusion}\label{sec:conclusion}

The proposed Aux-VAE method introduces a novel approach to variational autoencoder architecture, effectively integrating auxiliary (potentially non-exhaustive) information to enhance latent space disentanglement while preserving data generation quality. Demonstrated through extensive experiments across diverse datasets, Aux-VAE surpasses traditional VAEs and other disentanglement techniques, showcasing robustness and versatility due to its additional latents compensating for non-exhaustive auxiliary information, which is typical in scientific applications. 


Looking ahead, we will validate Aux-VAE on real-world datasets and deepen its theoretical foundations. In particular, we plan to move beyond our current polynomial, pairwise dependency measures by adopting mutual-information–based metrics that more efficiently capture nonlinear relationships between latent factors and auxiliary variables. We also intend to investigate potential downstream uses of Aux-VAE—for example, determining whether a candidate measurement truly contributes independent variation by training without it and then assessing its alignment with the learned latent dimensions. We believe these efforts will both refine Aux-VAE’s performance and broaden its applicability across diverse scientific domains.

\section*{Declarations}


\begin{itemize}
\item Funding: This work is supported by the U.S. Department of Energy, Office of Science, Office of Nuclear Physics, Office of Advanced Scientific Computing Research through the Scientific Discovery through Advanced Computing (SciDAC) program and through the FASTMath Institute, under contracts DE-AC02-06CH11357, DE-AC05-06OR23177, and DE-SC0023472, in collaboration with Argonne National Laboratory, Thomas Jefferson National Laboratory, National Renewable Energy Laboratory, and Virginia Tech. Work at Argonne National Laboratory was supported by the U.S. Department of Energy, Office of High Energy Physics. Argonne, a U.S. Department of Energy Office of Science Laboratory, is operated by UChicago Argonne LLC under contract no. DE-AC02-06CH11357. The training is carried out on Swing, a GPU system at the Laboratory Computing Resource Center (LCRC) of Argonne National Laboratory. This work was authored by the National Renewable Energy Laboratory, operated by Alliance for Sustainable Energy, LLC, for the U.S. Department of Energy (DOE) under Contract No. DE-AC36-08GO28308. 
\item Conflict of interest/Competing interests: The authors have no competing interests to declare that are relevant to the content of this article.
\item Ethics approval and consent to participate: Yes
\item Consent for publication: Yes
\item Data availability: We have provided the code in the supplementary material to reproduce the galaxy simulation data used in our experiments. Other used datasets are publicly available, we cited the original papers that introduced these datasets.
\item Materials availability: Not applicable
\item Code availability: The code for the proposed method
is available at \cite{AuxVAE2024} and further mentioned in the supplementary material. 
\item Author contribution: All authors contributed to the conception and design of the study. Methodology development and data analysis were conducted by Arkaprabha Ganguli, while Nesar Ramachandra generated the galaxy simulation dataset. Arkaprabha Ganguli drafted the initial version of the manuscript, and all authors provided critical feedback and revisions to earlier versions. All authors read and approved the final manuscript.
\end{itemize}

\noindent

\bigskip

\bibliography{sn-bibliography}


\newpage
\appendix
\section{Theoretical justification on regularizing the Expected Variational Posterior}
In this section, we formalize the problem setting and outline the underlying assumptions. Suppose that our true generative model is $x =g^*(s)+\epsilon$, where $\epsilon \sim N(0,\sigma^2)$ denotes the random fluctuations. We aim to learn a latent-variable model with prior $p(z)$ and generator $g$, where $g(z)\overset{d}{=} g^*(s)$. We also assume our access solely to a subset of the ground truth factors  $s_{obs} \subset s, s_{obs}\in \mathbb{R}^d$, conveyed through auxiliary variables $u \in \mathbb{R}^d$. Each auxiliary variable is intended to encapsulate a specific ground truth factor within $s_{obs}$. Hence, our observed database contains $n$ independently and identically distributed pairs of $x$ and $u$, denoted as $\mathcal{D} = \{(x^{(1)}, u^{(1)}), (x^{(2)}, u^{(2)}), \dots, (x^{(n)}, u^{(n)})\}$.  

\paragraph{Defining the VAE objective:} Under this setting, the generative process can be written as:
\begin{align}
    & z \sim p(z) \\
    & x \sim p_\theta(x|z,u) = p_\theta(x|z)
\end{align}
as, the latent factors $z$ collectively represents the whole ground-truth generative factors, conditional on $z$, $x$ and $u$ are independent. Similarly, to develop the VAE framework, our first pathway is the inference process, denoted $q_\phi(x,z,u)=q_\phi(z|x,u)q(x,u)=q_\phi(z|x)q(x,u)$. 
Now, to obtain a sample $(z,x,u)$ from this joint distribution, one would simply consider:
\begin{align}
    & x,u \sim q(x,u)\nonumber \\
    & z \sim q_\phi(z|x)
\end{align}
where $q(x,u)$ denotes the ground truth data distribution, and $q_\phi(z|x)$ is the learnable variational posterior. The inference process aims to extract latent representations from actual samples from the data distribution $q(x,u)$. Hence, as illustrated in \cite{VAE_intro}, one feasible approach to optimize wrt the KL distance: 
\begin{align}
    argmax_{\theta, \phi} & -KL\Bigl[q_\phi(x,z,u)||p_\theta(x,z,u)\Bigr] \nonumber\\
    & =\mathbb{E}_{q_\phi(x,z,u)}\Bigl[log\frac{p_\theta(x,z,u)}{q_\phi(x,z,u)}\Bigr]\nonumber\\
    & = \mathbb{E}_{q_\phi(z|x)}\bigl[log\frac{p_\theta(x|z,u) p(z,u)}{q_\phi(z|x)}\Bigr]-\mathbb{E}_{q_\phi(z|x)}log q(x,u))\nonumber\\
    & = \mathbb{E}_{q_\phi(z|x)}\bigl[log p_\theta(x|z)\Bigr] + \mathbb{E}_{q_\phi(z|x)}\bigl[log\frac{p(z,u)}{q_\phi(z|x)}\Bigr] -\text{constant} \nonumber\\
    & = \mathbb{E}_{q_\phi(z|x)}\bigl[log p_\theta(x|z)\Bigr] + \mathbb{E}_{q_\phi(z|x)}\bigl[log\frac{p(z|u)}{q_\phi(z|x)}\Bigr] -\text{constant} \nonumber\\
    & = \mathbb{E}_{q_\phi(z|x)}\bigl[log p_\theta(x|z)\Bigr]  - KL \Bigl[q_\phi(z|x)||p(z|u)\Bigr] -\text{constant}\nonumber\\
    & = \text{likelihood} - KL \Bigl[q_\phi(z|x)||p(z|u)\Bigr] -\text{constant}
\end{align}

\paragraph{Regularization over Expected Variational Posterior:} 
As discussed in Section 2 of the main manuscript, enforcing disentanglement can be approached by directly minimizing the KL divergence between the expected variational posterior $q_\phi(z) = \int q_\phi(z|x)p(x)dx$ and the prior $p(z)$. However, this KL term lacks a closed-form expression, which complicates optimization efforts. As an alternative, we propose to promote the major structural properties of $p(z)$ within $q_\phi(z)$. Specifically, we concentrate on the following three disentangled properties of $p(z)$ which can be expressed through its conditional distribution wrt $u$:
\begin{itemize}
    \item \textbf{\textit{Inter-independence}: $u \perp z_{recon}$}
    \paragraph{Proof:} As we observe that, for $1\leq j\leq d, 1 \leq j^{'} \leq d_Z-d$, 
    \begin{align}
        \mathbb{E}(u_j  z_{j^{'}}) &= \int u_j z_{j^{'}}p(u_j, z_{j^{'}})du_j dz_{j^{'}} \nonumber\\
        & =\int u_j z_{j^{'}} p(z_{j^{'}}|u_j )p(u_j )du_j dz_{j^{'}} \nonumber\\
        & = \int u_j z_{j^{'}} p(z_{j^{'}})p(u_j )du_j dz_{j^{'}} \nonumber\\
        & = \mathbb{E}(u_j)\mathbb{E}(z_{j^{'}})
    \end{align}
    \item \textbf{\textit{Intra-independence}:} $u_{j} \perp z_{aux,j'}$ for $j,j'=1,2,\dots d, j \neq j'$
    \paragraph{Proof:} 
    \begin{align}
        \mathbb{E}(u_j z_{j^{'}}) &= \int u_j z_{j^{'}}p(u_j, z_{j^{'}})du_j dz_{j^{'}} \nonumber\\
        & =\int u_j z_{j^{'}} p(z_{j^{'}}|u_j )p(u_j )du_j dz_{j^{'}} \nonumber\\
        & = \int u_j z_{j^{'}} p(z_{j^{'}})p(u_j )du_j dz_{j^{'}} \nonumber\\
        & = \mathbb{E}(u_j)\mathbb{E}(z_{j^{'}})
    \end{align}
    \item \textbf{\textit{Explicitness}:} $Corr(u_j,z_j) \to 1$, for $1\leq j\leq d$ as $n\to \infty$.
    \paragraph{Proof:} 
    By the theorem of total probability, 
    
    \begin{align*}
        & cov(u_j,z_j)=E_u cov_{u,z \sim p(z|u)}(u_j, z_j) + cov_u(u_j, E(z_j|u)) = cov(u_j, u_j)=var(u_j)\\
        & var(z_j)= E_u var(z_j|u)+var_u(E(z_j|u))=\frac{1}{n}+var(u_j)\\
        & corr(u_j,z_j) = \frac{cov(u_j,z_j)}{\sqrt{var(u_j) var(z_j)}} = \frac{var(u_j}{\sqrt{(\frac{1}{n}+var(u_j))var(u_j)}} \to \infty
    \end{align*}
    as $n \to \infty$
\end{itemize}
Quantitatively measuring these non-linear dependencies to measure Inter and Intra-independence strength under $q_\phi(z)$ is also non-trivial, and we turn our attention to polynomial regression here. 
\begin{itemize}
    \item \textbf{\textit{Inter-independence}: } To measure the Inter-independence strength, we aggregate the correlation between different degrees of polynomials of $u$ and $Z_{recon}$. e.g. 
    \begin{align}\label{eq:total_variance}
    Cov(u^k, z_{recon}^{k'})= & E_{(x,u) \sim p(x,u)}Cov_{u, z \sim q_\phi (z|x)}(u^k, z_{recon}^{k'}) + \nonumber\\ 
    &Cov_{(x,u)} (u^k, E_{z \sim q_\phi(z/x)}(z_{recon}^{k'}) \nonumber\\
& = Cov_{(x,u)} (u^k, \mu_{\phi, d+1:d_Z}^{k'}).
\end{align}
Hence, we simply use a running estimate of these correlations between the u and the means of the latent factors to create the summary statistics $R_0(\cdot, \cdot)$ and $R_1(\cdot, \cdot)$which are informative in assessing the non-linear dependency. Similarly, 
\item \textbf{\textit{Intra-independence}:}
\begin{align}\label{eq:total_variance_1}
    Cov(u_j^k, z_{aux,j^{'}}^{k'})= & E_{(x,u) \sim p(x,u)}Cov_{u, z \sim q_\phi (z|x)}(u_j^k, z_{aux,j^{'}}^{k'}) + \nonumber\\ 
    &Cov_{(x,u)} (u_j^k, E_{z \sim q_\phi(z|x)}(z_{aux,j^{'}}^{k'}) \nonumber\\
& = Cov_{(x,u)} (u_j^k, \mu_{\phi, j^{'}}^{k'}), \text{ for } j,j'=1,2,\dots d, j \neq j'.
\end{align}
and  
\item \textbf{\textit{Explicitness}:} This property also indicates that under $q_\phi(z|u)$, the correlation between $Z_{aux,j}$ and $u_j, j=1,2,\dots, d$ should be strong. Hence, we calculate 

\begin{align}\label{eq:total_variance_2}
    Cov(u_j, z_{aux,j})= & E_{(x,u) \sim p(x,u)}Cov_{u, z \sim q_\phi (z|x)}(u_j, z_{aux,j}) + \nonumber\\ 
    &Cov_{(x,u)} (u_j, E_{z \sim q_\phi(z|x)}(z_{aux,j}) \nonumber\\
& = Cov_{(x,u)} (u_j, \mu_{\phi, j}), \text{ for } j=1,2,\dots d.
\end{align}
Hence, 
\begin{align}\label{eq:total_variance_3}
    Corr(u_j, z_{aux,j}) & =  \frac{Cov(u_j, z_{aux,j})}{\sqrt{var(u_j) var(z_{aux,j})}} \nonumber\\ &= \frac{Cov(u_j, \mu_{\phi, j})}{\sqrt{var(u_j) (E(var_{z \sim q_\phi(z|x)}(z_{aux,j}))+var(\mu_{\phi, j})}}
\end{align}
\end{itemize}
This incorporates $Corr(u_j, z_{aux,j})<Corr_{(x,u)} (u_j, \mu_{\phi, j}), \text{ for } j=1,2,\dots d.$ While it is theoretically feasible to regularizing $Corr(u_j, z_{aux,j})$ in eq. \ref{eq:total_variance_3} towards one, we observed through our experiments that regularizing $Corr_{(x,u)} (u_j, \mu_{\phi, j})$  is computationally much more efficient and achieves similar levels of accuracy. For datasets with higher complexity, however, one may need to directly regularize  $Corr(u_j, z_{aux,j})$ using eq. \ref{eq:total_variance_3} using eq. \ref{eq:total_variance_3} to capture the finer relationships within the data. 

Hence, the final objective function of Aux-VAE incorporates the concepts of non-linear dependency between the latent factors and the auxiliary information into the optimization of the ELBO, thus effectively achieving the desired disentanglement. When constructing the main loss for Aux-VAE, we focus on Pearson's correlation coefficient due to its bounded nature and use running estimates over the mini-batch for the correlation.

\section{Additional Experimental Details}
\subsection{Brief description of the galaxy simulation data}\label{sup:galaxies}

In simulated and experimental datasets in scientific research, a subset of the auxiliary variables
\begin{table}[h]
    \centering
    \caption{Parameter descriptions and ranges of galaxy image dataset}
    \label{tab:parameters}
    \begin{tabular}{|l|p{8cm}|l|}
        \hline
        \textbf{Parameters} & \textbf{Short description} & \textbf{Range} \\ \hline
        \multirow{2}{*}{$flux$} & \multirow{2}{8cm}{Apparent brightness of the galaxy (Number counts)} & \multirow{2}{*}{$10^4$ -- $10^5$} \\ 
         &  &  \\ \hline
        \multirow{2}{*}{$radius$} & \multirow{2}{8cm}{Radius of the galaxy (Arc-seconds)} & \multirow{2}{*}{0.1 -- 1} \\ 
         &  &  \\ \hline
        \multirow{2}{*}{$g1$, $g2$} & \multirow{2}{8cm}{Reduced gravitational shear components (Cartesian coordinates)} & \multirow{2}{*}{-0.5 -- 0.5} \\ 
         &  &  \\ \hline
        \multirow{2}{*}{$psf$} & \multirow{2}{8cm}{Full width at half maximum of the point-spread function} & \multirow{2}{*}{0.2 -- 0.4} \\ 
         &  &  \\ \hline
    \end{tabular}
\end{table}
 may be `controlled’ in the experimental design. Whereas in observational research domains such as astronomy, such quantities may just be measured or inferred either directly or with complementary studies. The dataset we have utilized in this effort is a representative simulation of telescopic observations created using GalSim \cite{Galsim2015}. We assume that the light profile of each galaxy can be approximated as an exponential disk, which is known to be a good description of the outer, star-forming regions of spiral galaxies \cite{Lacker2012}. In reality, galaxies can exhibit spiral, elliptical, barred spirals, irregulars, and other diverse morphologies \cite{hubble2015}. 
Telescopes are often systematically sensitive to certain types of galaxies, depending on which stage of galaxy evolution is probed based on instrumental specifications. For creating the synthetic galaxy image dataset, we first consider 5 physical parameters of varying importance. Ranges of these parameters are heuristically determined based on real observations, a Latin-Hypercube sampling over the range is performed to select 16,384 simulation points. The details of the ranges of the parameters are shown in Table \ref{tab:parameters}. Each galaxy image is created with 33x33 pixels, with galaxies at the centers. We also note that while the galaxies here are in grayscale, the majority of the modern telescopes observe the Universe in multiple bands of channels of the light spectrum.

\subsubsection{Relative importance between $Z_{aux}$ and $Z_{recon}$}
\begin{figure}
      \centering
     \includegraphics[height=10cm,width=10cm]{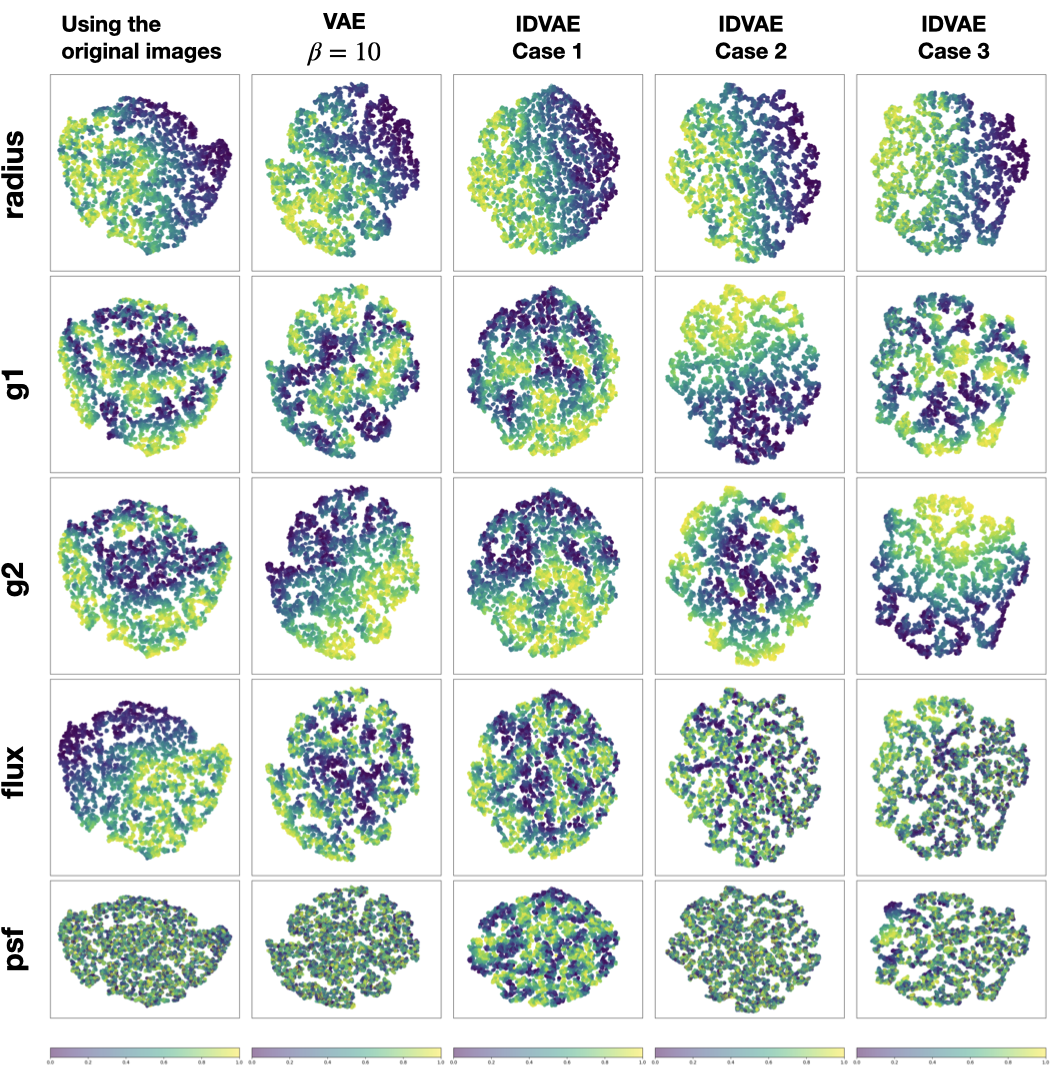}
      
      \caption{Comparative t-SNE Visualization of Latent Spaces from VAE and IDVAE Across Multiple Scenarios with Respect to Generative Factors for the Galaxy Simulation Dataset}
      \label{fig:t-sne_others}
\end{figure}
\begin{figure}
      \centering
     \includegraphics[height=10cm,width=13.5cm]{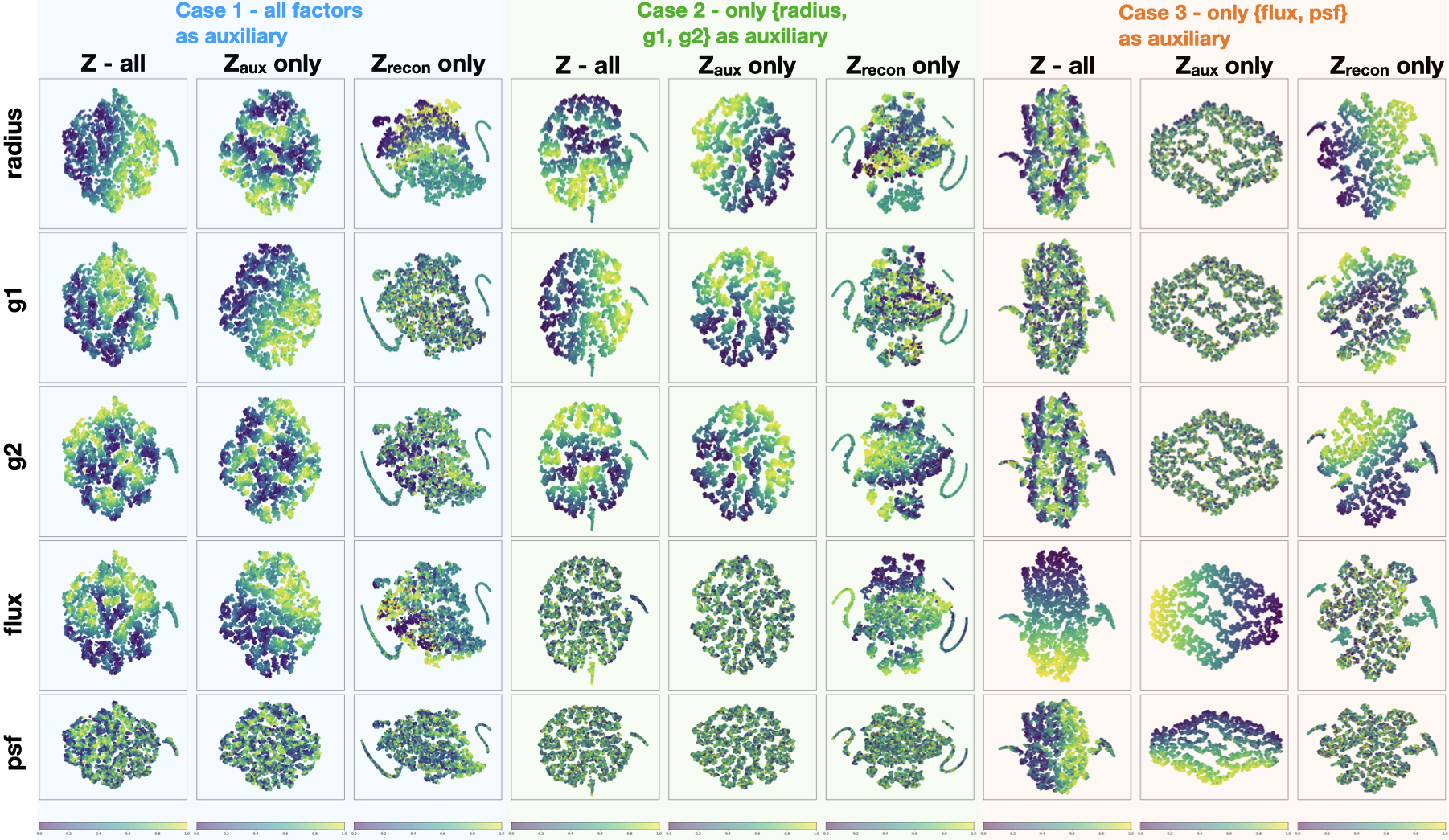}
      
      \caption{Distinct Impact of $Z_{aux}$ and $Z_{recon}$ on Generative Factor Representation in Aux-VAE's Latent Space: A t-SNE Visualization}
      \label{fig:t-sne_ours}
\end{figure}
To elucidate the roles of $Z_{aux}$ and $Z_{recon}$, we conducted a t-SNE analysis \citep{t-sne}, illustrated in Figures \ref{fig:t-sne_others} and \ref{fig:t-sne_ours}. This analysis visualized the 2D components of t-SNE representations derived from original test images and latent factors from VAE, IDVAE, and Aux-VAE models across three scenarios outlined in Section 3 of the main manuscript. The plots are organized in a grid, with each column representing different configurations and each row color-coded by one of five generative factors, providing a method to assess each model's factor representation.

\paragraph{Key Observations:}
\begin{enumerate}
    \item \textbf{From t-SNE representation of latent spaces of competing methods in Figure \ref{fig:t-sne_others}}:
    \begin{itemize}
        \item The first column, featuring t-SNE plots of original images, reveals the minimal impact of the $psf$ factor on generative modeling, as shown by the absence of distinct clustering for $psf$.
        \item The second column displays t-SNE plots of VAE's 10-dimensional latent factors from 1000 test images, indicating VAE's ability to recognize underlying generative factors, albeit with entangled representations. This suggests that while VAE identifies different factors, it has difficulty clearly separating them in the latent space.
        \item Columns 3-6 present t-SNE plots from IDVAE's application on cases 1, 2, and 3 (detailed in Section 3). Despite incorporating $flux$ and $psf$ as auxiliary information in Case 3, IDVAE did not effectively represent these factors, struggling to distinctly segregate them in the latent space, which points to shortcomings in its auxiliary data integration.
    \end{itemize}
    
    \item \textbf{From the t-SNE representation of latent space of Aux-VAE in Figure \ref{fig:t-sne_ours}}: This analysis was segmented into three parts, each examining the t-SNE representations of all latent factors, as well as the separate contributions of $Z_{aux}$ and $Z_{recon}$ to the preservation of generative factor information in the latent space.
    \begin{itemize}
    
        \item \textbf{Case 1:} Here, all generative factors are encapsulated by the auxiliary features, and thus by $Z_{aux}$. The t-SNE plots show similar patterns for $Z-$all' and $Z_{aux}$ only', indicating that $Z_{recon}$ does not significantly contribute additional information regarding the generative factors in this scenario.
        \item \textbf{Cases 2 and 3:} A consistent pattern emerges across these cases. The generative factors that $Z_{aux}$ covers are clearly depicted by the corresponding latent factors, while $Z_{recon}$ effectively captures the remaining factors. For instance, in case 3, factors like $flux$, $g1$, and $g2$ are distinctly represented in the t-SNE plots of $Z_{recon}$, demonstrating its effectiveness in portraying uncovered generative aspects.
    \end{itemize}
\end{enumerate}
This analysis underscores how Aux-VAE dynamically adjusts the relative importance of $Z_{aux}$ and $Z_{recon}$, from scenarios with comprehensive auxiliary information (case 1) to those with limited auxiliary data (case 3). This flexibility demonstrates Aux-VAE's capability to adapt and effectively utilize the available information to maintain accurate representation of underlying generative factors.

\subsubsection{Adversarial robustness}
\begin{figure}[!tbp]
  \centering
    \includegraphics[width=\textwidth]{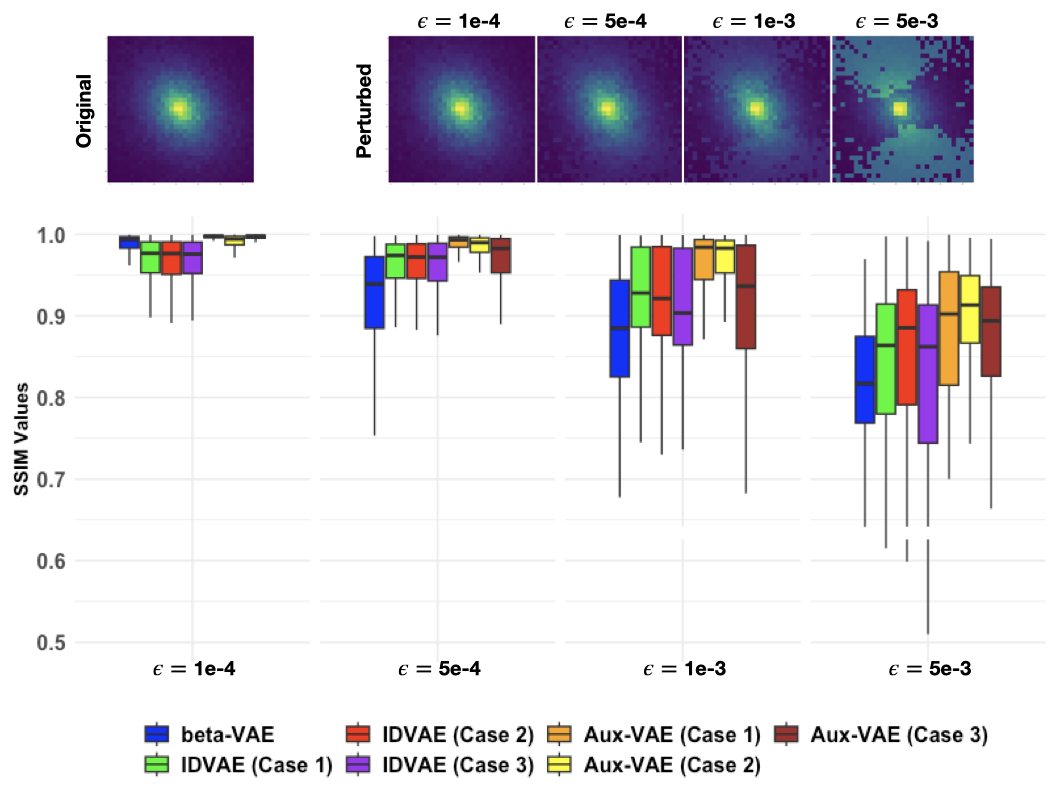}
    \caption{Perturbed images and SSIM boxplots evaluating adversarial robustness across Beta-VAE, IDVAE, and Aux-VAE models under FGSM attack.}
    \label{fig:advarsarial}
\end{figure}
In the adversarial robustness comparison experiment, we subjected the models to a Fast Gradient Sign Method (FGSM) attack, a common technique for testing model robustness against adversarial 
examples \citep{advarsarial_fsgm}. For an input image, the FSGM method uses the gradients of the loss with respect to the input image to create a perturbed new image that maximizes the loss. The perturbation strength is controlled by a parameter epsilon ($\epsilon$). These perturbed images were then used to evaluate the models' performance under attack. In Figure \ref{fig:advarsarial}, an illustration of the perturbed images under varying perturbation strength $\epsilon$ is presented. Additionally, for a test set of 1000 images, we calculate the SSIM metric between the input images and the reconstructed images after the FGSM attack, and the boxplots are presented in Figure \ref{fig:advarsarial}. Naturally, we see a decline in SSIM-metric with the increase in perturbation strength $\epsilon$. As anticipated, $\beta$-VAE, lacking proper disentanglement, exhibited heightened vulnerability to the adversarial attack. Conversely, both IDVAE and Aux-VAE, which demonstrate some level of disentanglement, exhibited comparatively greater robustness. Notably, in Case 3, where the auxiliary information lacks representation of several crucial generative factors, 
both IDVAE and Aux-VAE displayed relatively poor performance, suggesting the importance of comprehensive auxiliary information for enhanced adversarial robustness.

\subsubsection{Disentanglement among the latent factors wrt the ground-truth generating factors - On the galaxy simulation data, Cars3D and Dsprites datasets}
Following Section 3 of the main manuscript, here we present the remaining results on Disentanglement among the latent factors wrt the ground-truth generating factors across VAE, IDVAE and Aux-VAE.
\begin{figure}[!tbp]
  \centering
    \includegraphics[height=6cm,width=10cm]{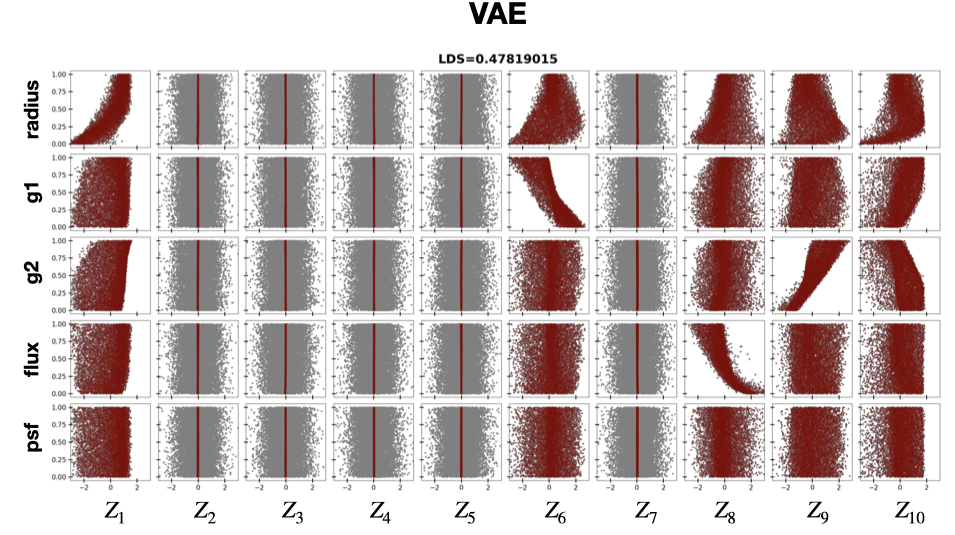}
  \caption{Visualizing Disentanglement for VAE: This scatterplot contrasts latent factors ($Z$, represented by grey dots) with the latent means ($\mu_\phi$, shown as maroon dots), alongside highlighting the LDS metric. }
  \label{fig:z-vs-u-VAE}
\end{figure}

\begin{figure}[!tbp]
  \centering
    \includegraphics[height=8cm,width=14cm]{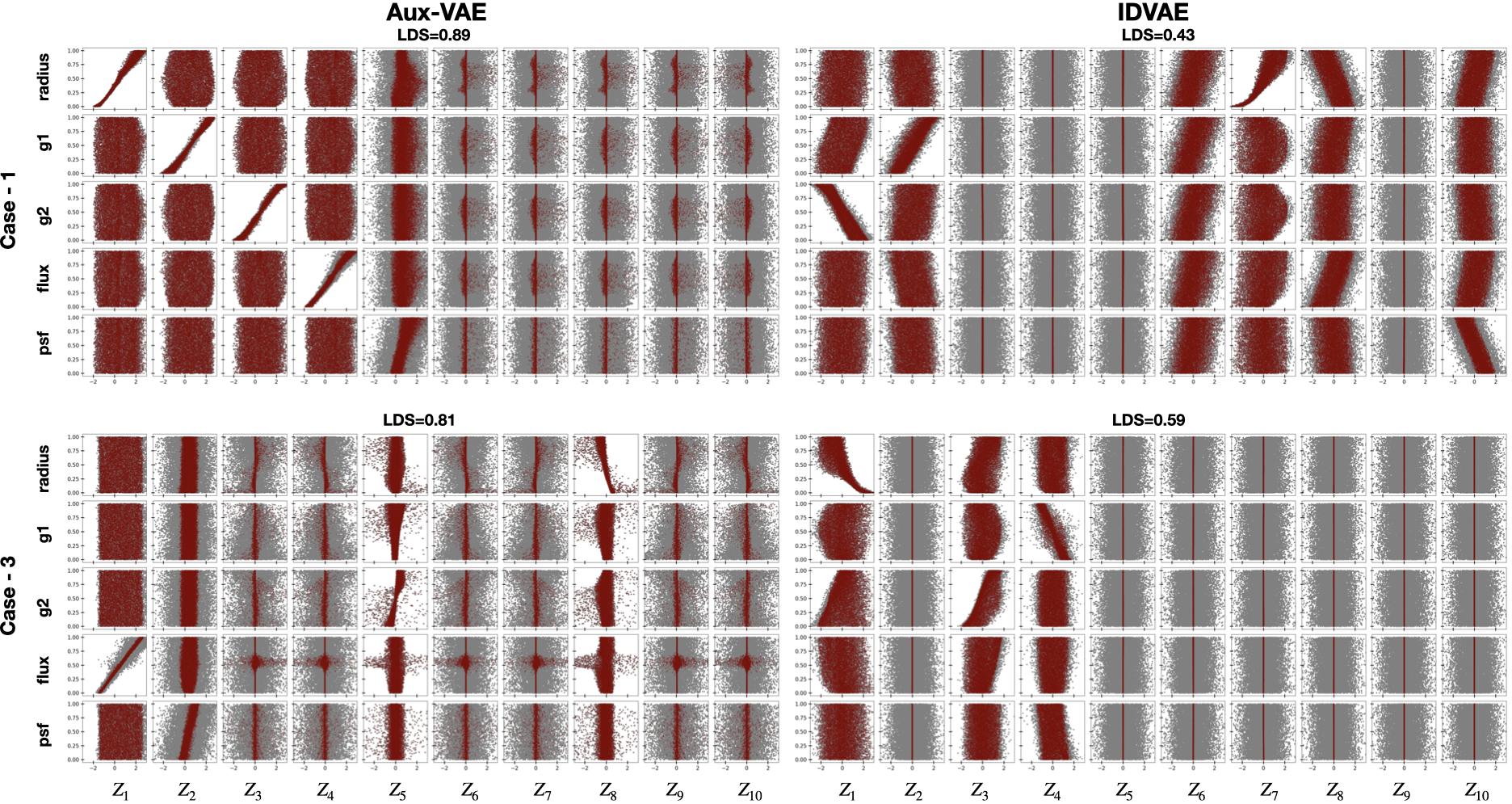}
  \caption{Disentanglement Visualization for IDVAE and Aux-VAE: This scatterplot illustrates the comparison between latent factors ($Z$, depicted as grey dots) and latent means ($\mu_\phi$, represented as maroon dots), with an emphasis on the LDS metric. The plot includes results for cases 1 and 3 of the galaxy simulation dataset, with case 2 detailed in Section 3 of the main manuscript.}
  \label{fig:z-vs-u-IDVAE_AuxVAE}
\end{figure}


\begin{figure}[!tbp]
  \centering
    \includegraphics[height=9cm,width=15cm]{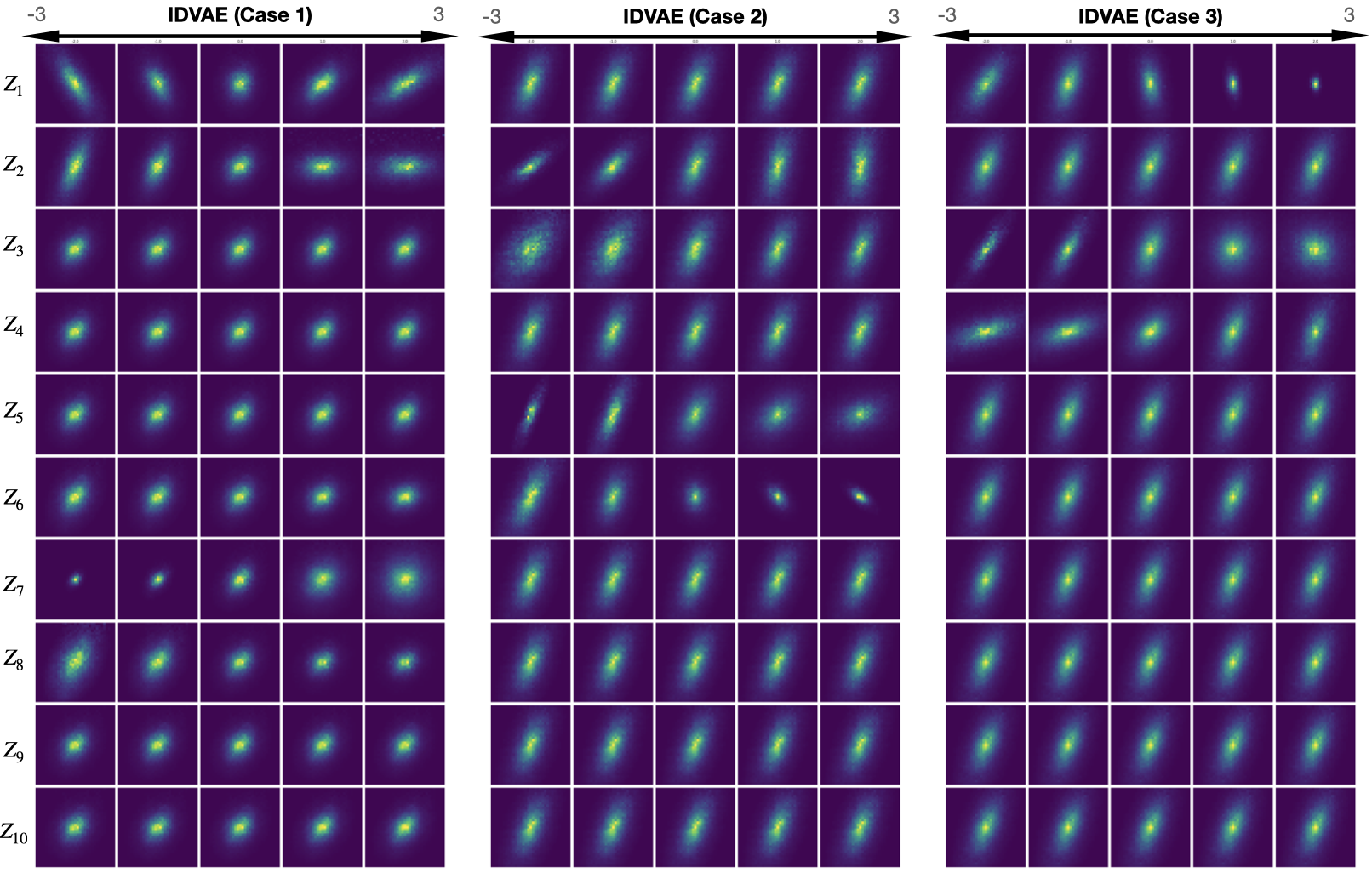}
  \caption{Latent space traversal for IDVAE on the three cases considered for the galaxy simulation dataset.  }
  \label{fig:latent_travarsal_IDVAE_all}
\end{figure}

\begin{figure}[!tbp]
  \centering
    \includegraphics[height=9cm,width=15cm]{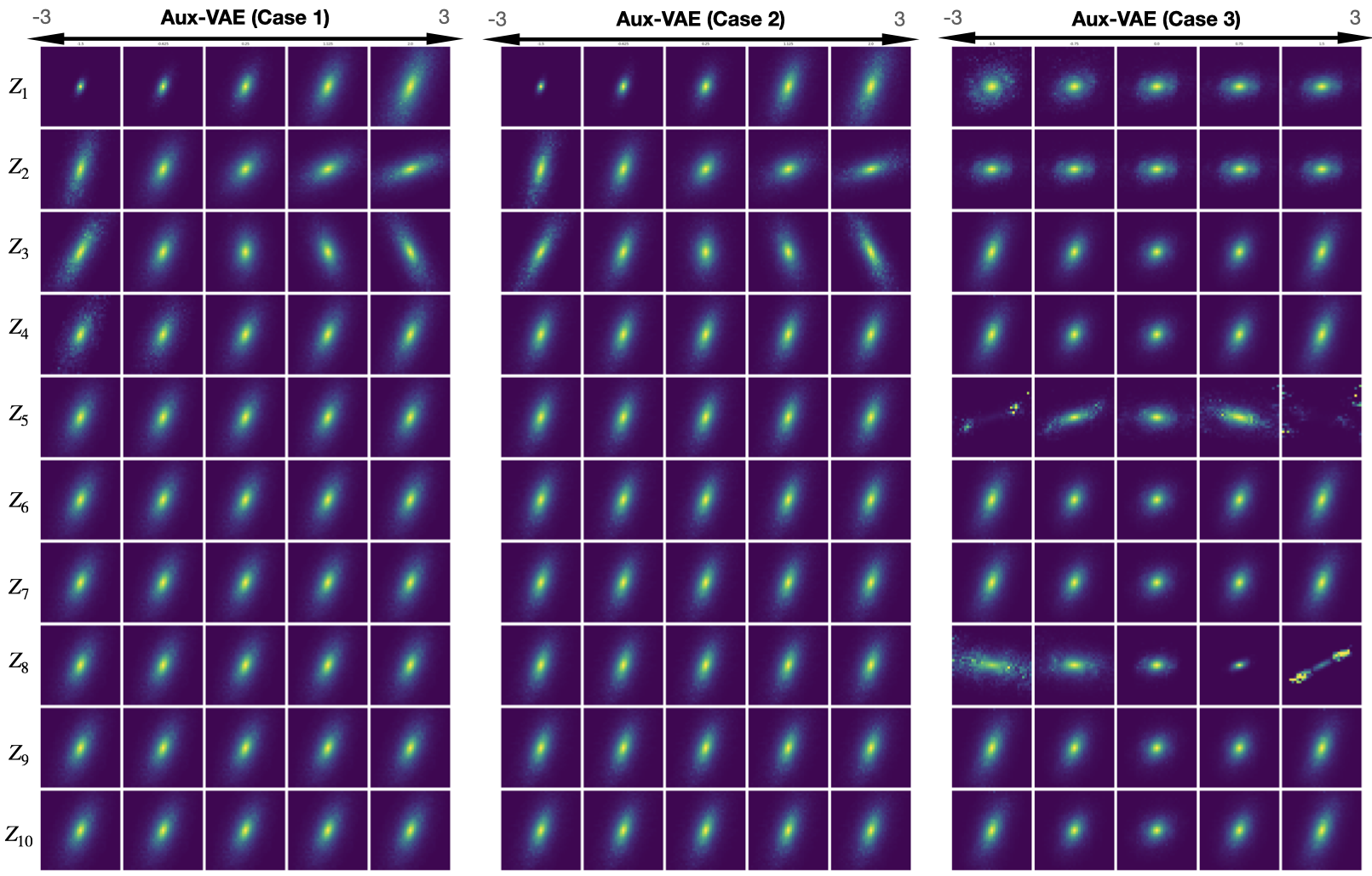}
  \caption{Latent Space Exploration for Aux-VAE Across Three Scenarios in the Galaxy Simulation Dataset. In case 1, $Z_{aux}$ includes $Z_{1:5}$; in case 2, $Z_{aux}$ comprises $Z_{1:3}$; and in case 3, $Z_{aux}$ consists of $Z_{1,2}$. The results demonstrate that in each scenario, $Z_{aux}$ effectively adapts to the associated generative factors. }
  \label{fig:latent_travarsal_ours_all}
\end{figure}

\begin{figure}[!ht]
\centering
\begin{subfigure}{\textwidth}
  \centering
  \includegraphics[height=7cm,width=9cm]{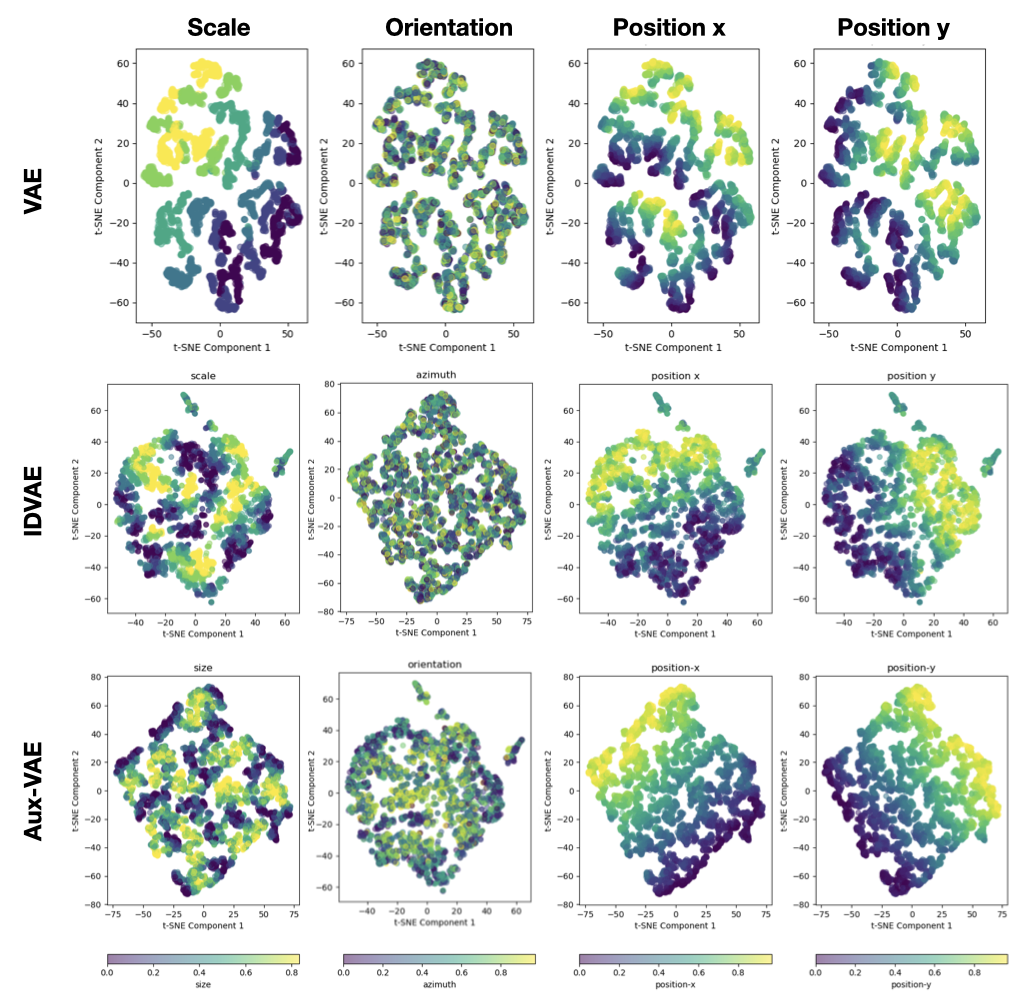}
  \caption{Latent space representation on \textbf{DSprites} dataset by t-SNE: four columns represent the generative factors 'Scale', 'Orientation', `Position x', and `Position y'. Aux-VAE achieves better disentanglement compared to the other competing methods. }
  \label{fig:t-sne_DSprites}
\end{subfigure}%
\vspace{1cm} 
\begin{subfigure}{\textwidth}
  \centering
  \includegraphics[height=8cm,width=8cm]{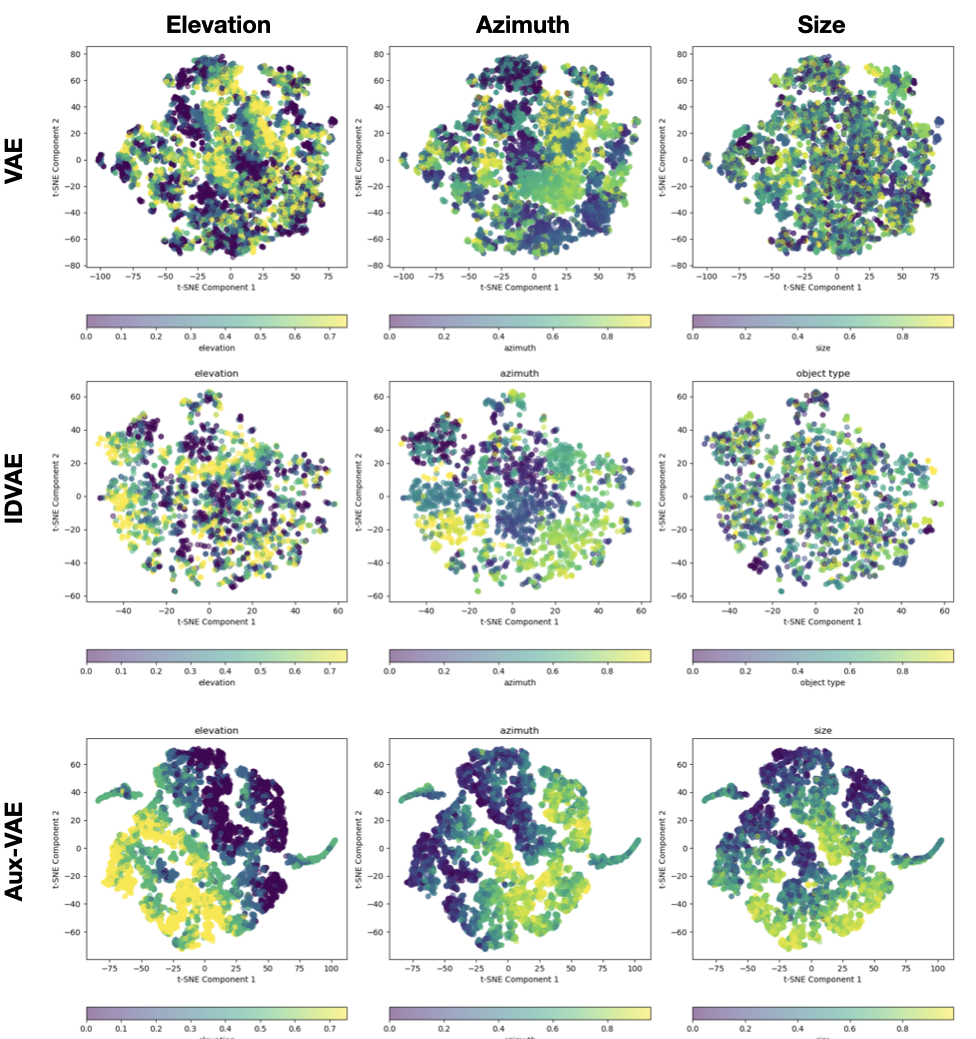}
  \caption{Latent space representation on \textbf{Cars3D} dataset by t-SNE: three columns represent three generative factors `height', `azimuth', and `size'. Aux-VAE achieves better disentanglement than the other competing methods. }
  \label{fig:t-sne_Cars3D}
\end{subfigure}
\caption{Latent space representation on \textbf{Cars3D} and \textbf{DSprites} dataset by t-SNE}
\label{fig:test}
\end{figure}
\begin{figure}[!ht]
\centering
\begin{subfigure}{\textwidth}
  \centering
  \includegraphics[height=5.5cm,width=15cm]{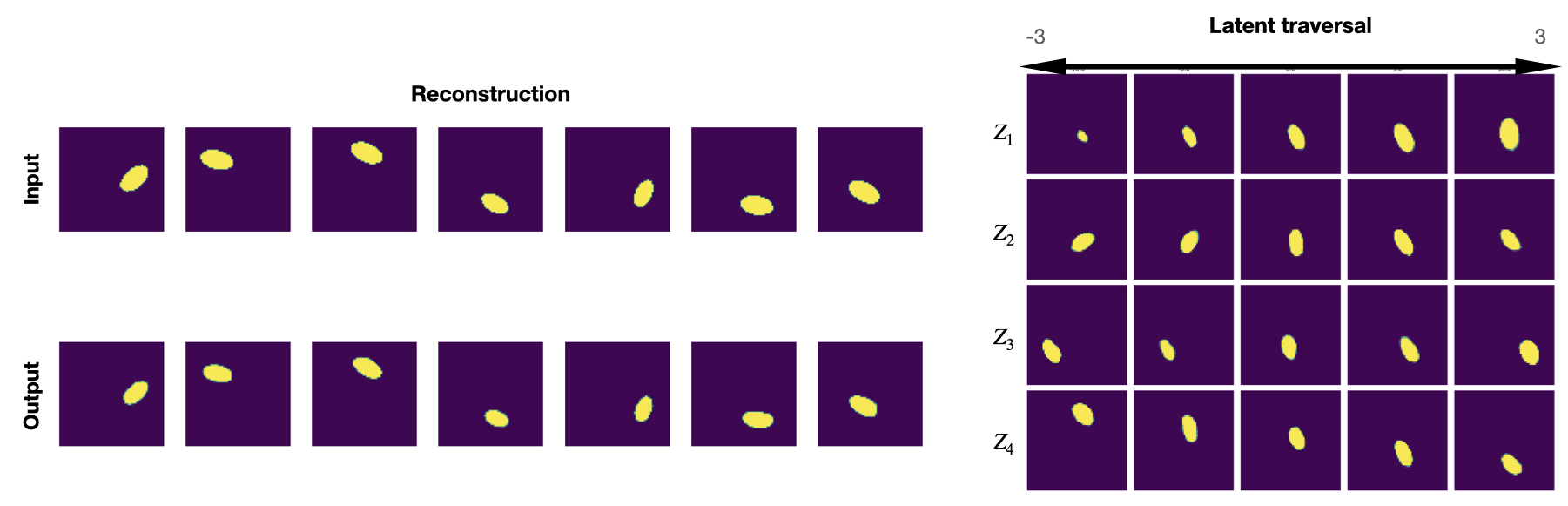}
  \caption{Evaluation of Aux-VAE on the \textbf{DSprites} Dataset: Four columns display the generative factors 'Scale', 'Orientation', 'Position x', and 'Position y', demonstrating the model's reconstruction accuracy and latent space traversal. This analysis focuses exclusively on circular shapes within the dataset. Aux-VAE exhibits superior disentanglement relative to competing models, aligning with findings from previous studies \cite{beta-TCVAE, IDVAE}. Notably, Aux-VAE shows enhanced precision in capturing the 'Orientation' factor, a challenge where other models have shown limitations.}
  \label{fig:latent_traversal_DSprites}
\end{subfigure}%
\vspace{1cm} 
\begin{subfigure}{\textwidth}
  \centering
  \includegraphics[height=7cm,width=15cm]{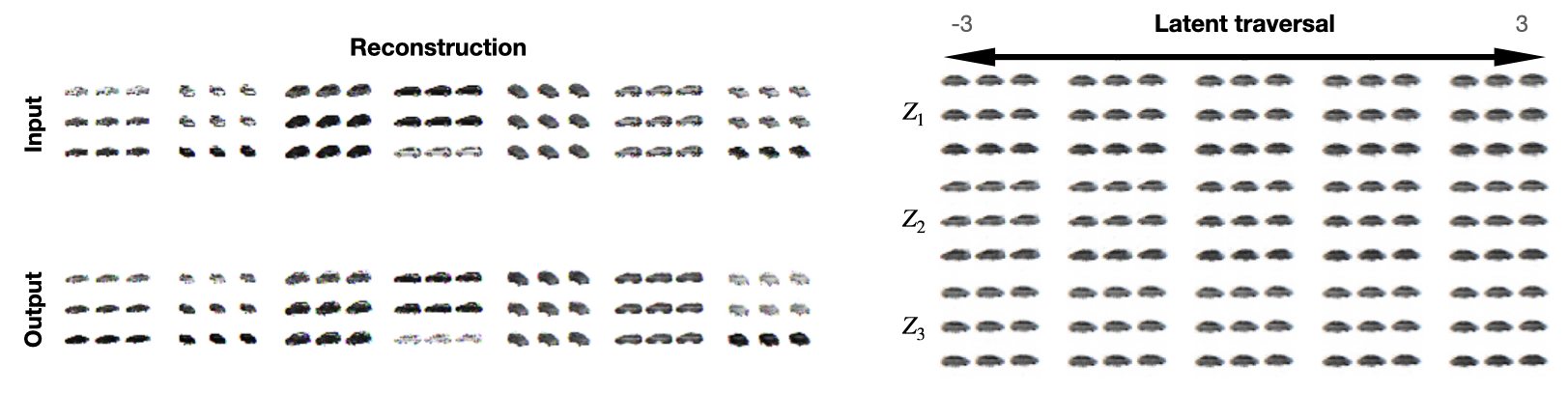}
  \caption{Demonstration of Aux-VAE's reconstruction accuracy and latent space traversal on \textbf{Cars3D} dataset: four columns represent the generative factors `Scale', `Orientation', `Position x', and `Position y'. Aux-VAE achieves better disentanglement compared to the other competing methods. }
  \label{fig:latent_traversal_Cars3D}
\end{subfigure}
\caption{Demonstration of Aux-VAE's reconstruction accuracy and latent space traversal on \textbf{DSprites} and \textbf{Cars3D} dataset}
\label{fig:test}
\end{figure}
\newpage

\begin{figure}[!ht]
\centering
\begin{subfigure}{\textwidth}
  \centering
  \includegraphics[height=5.5cm,width=14cm]{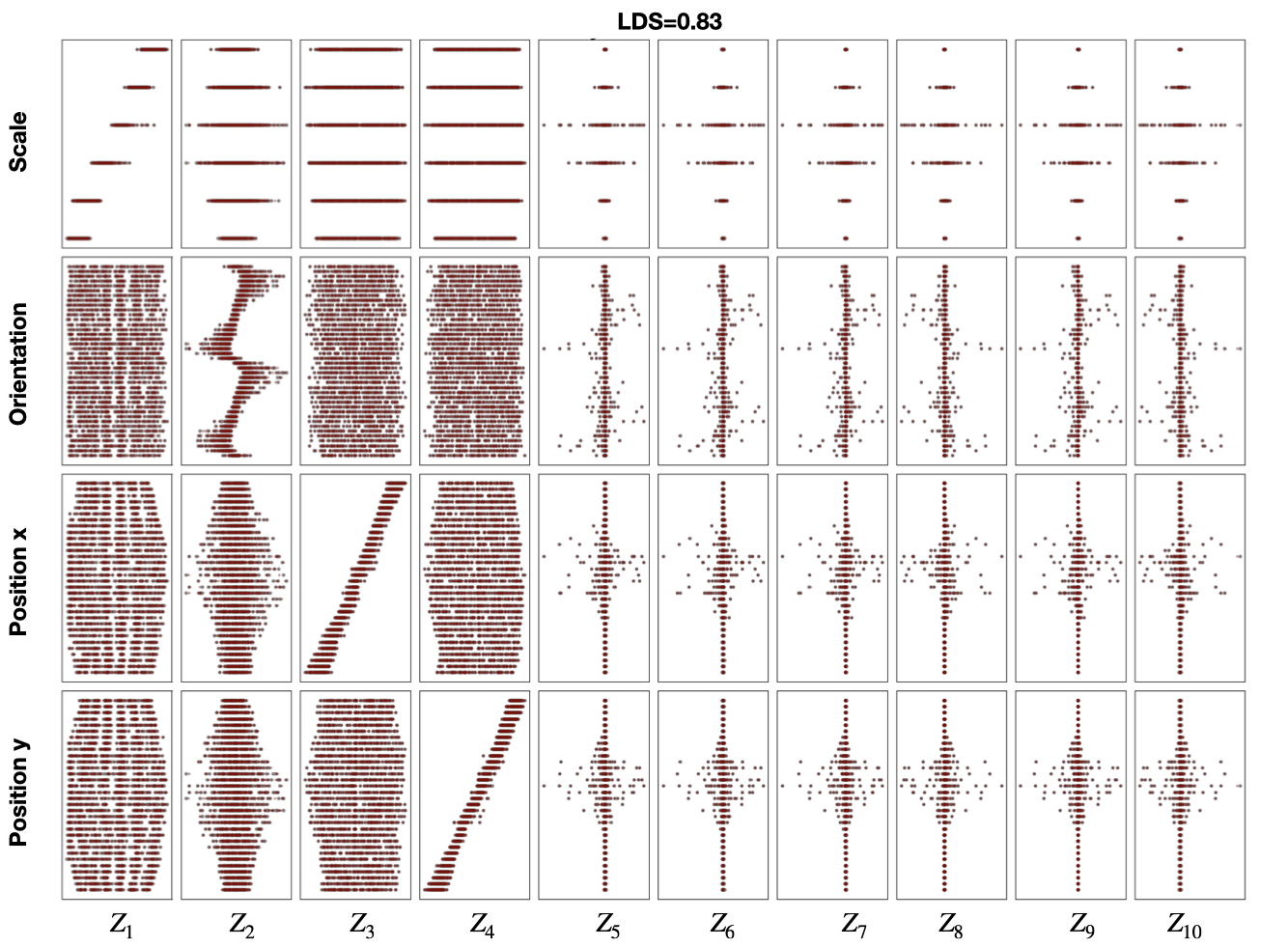}
  \caption{Visualizing Disentanglement in \textbf{`DSprites'} dataset: This scatterplot contrasts latent factors ($Z$, represented by grey dots) with the latent means ($\mu_\phi$, shown as maroon dots), alongside highlighting the LDS metric.  }
  \label{fig:Z_VS_U_DSprites}
\end{subfigure}%
\vspace{1cm} 
\begin{subfigure}{\textwidth}
  \centering
  \includegraphics[height=5.5cm,width=15cm]{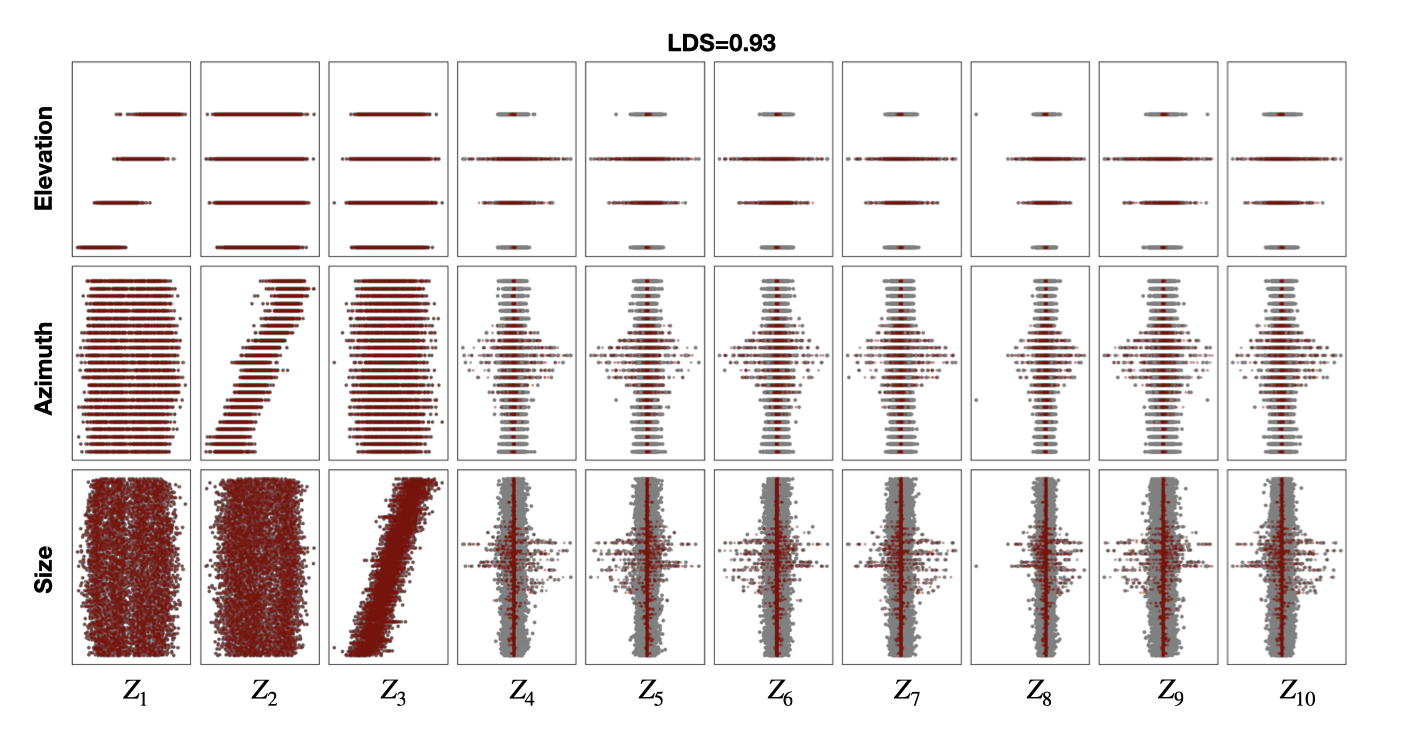}
  \caption{Visualizing Disentanglement in \textbf{`Cars3D'} dataset: This scatterplot contrasts latent factors ($Z$, represented by grey dots) with the latent means ($\mu_\phi$, shown as maroon dots), alongside highlighting the LDS metric. }
  \label{fig:Z_VS_U_Cars3D}
\end{subfigure}
\caption{Disentanglement in  \textbf{`DSprites'} and \textbf{`Cars3D'} dataset: Each latent factor is plotted against the underlying generative factors available. It shows the first $d$ latent factors (where $d$= no. of generative factors) are properly disentangled wrt the corresponding generative factors. }
\label{fig:Z_VS_U_all}
\end{figure}
\clearpage
\section{Details on model evaluation, hyperparameter tuning, and code repository}
\subsection{ Architectures}
\begin{figure}[!tbp]
  \centering
    \includegraphics[height=10cm,width=14cm]{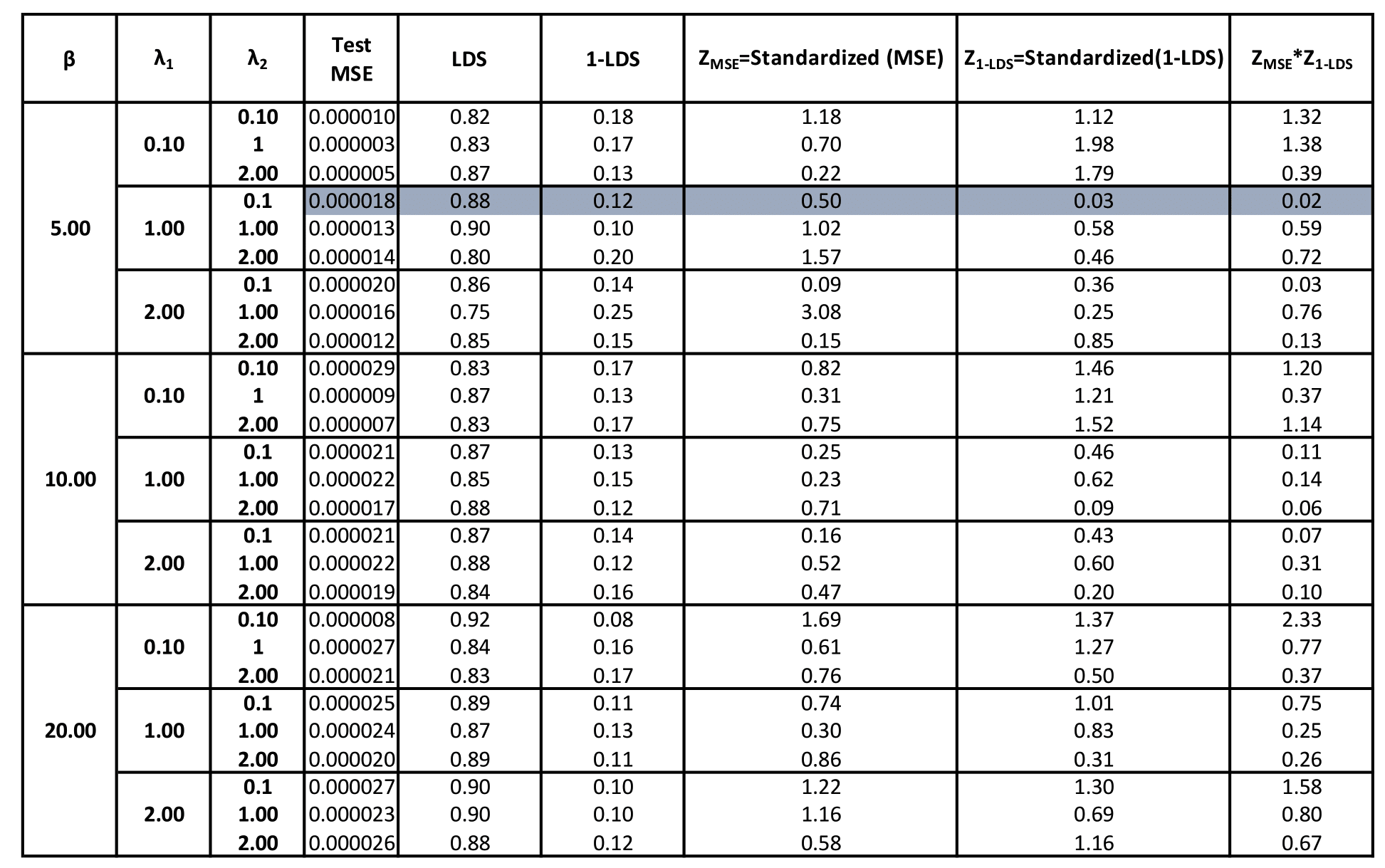}
  \caption{Hyperparameter Tuning for Aux-VAE on Galaxy Simulation Data: Grid Search Approach. The final scores are rounded up to two decimal places. The selected configuration, highlighted in the table, effectively balancethe s optimization of MSE and disentanglement in the latent factors.}
  \label{fig:hyperparameter-tuning-AuxVAE}
\end{figure}

The architecture for the competing methods, $\beta-$VAE and IDVAE, follows the specifications from their respective repositories: \href{https://github.com/matthew-liu/beta-vae.git}{$\beta-$VAE-repo} and \href{https://github.com/grazianomita/disentanglement_idvae.git}{IDVAE repo}. Each of the training is carried out on a single-NVIDIA A100 GPU. For Aux-VAE (code available at \cite{AuxVAE2024}
), we employed a basic grid-search approach for hyperparameter tuning, evaluating the mean squared error (MSE) and the disentanglement score (LDS) across different configurations. Specifically, we split the data into 7:2:1 as train, validation and test split. On the validation split, we tested $d_r$ different combinations of hyperparameters $(\beta, \lambda_1, \lambda_2)$, calculating the test MSE and test LDS for each configuration, denoted as $MSE_1, MSE_2,\dots, MSE_{d_r}$ and $LDS_1, LDS_2,\dots, LDS_{d_r}$. We standardized these criteria and calculated their product $MSE(1-LDS)$ to identify the optimal hyperparameter setting that jointly optimizes both reconstruction and disentanglement. Figure \ref{fig:hyperparameter-tuning-AuxVAE} illustrates the experiment and the hyperparameter values to select for the experiment. Table \ref{tab:auxvae-hyperparams} shows the final selected values of the hyperparameters for each dataset. For other architecture details, we set batch size=64, and learning rate=1e-3 with Adam as the optimizer. However, we recognize that with an increasing number of hyperparameters, grid-search becomes impractical. In such cases, one would adopt stochastic hyperparameter optimization algorithms, such as Deep-hyper \citep{deephyper_software, ytopt}, for a more efficient search. Table \ref{tab:beta_vae_cnn_aux} outlines the basic configuration of Aux-VAE used for the galaxy simulation dataset. For the Cars3D and DSprites datasets, the configurations are adjusted to align with the IDVAE settings discussed in \cite{IDVAE}. 

For $\beta-$VAE and IDVAE, we checked the MSE and LDS scores to find the optimal level of regularization, in the case of galaxy simulation data analysis.  For our \(\beta\)-VAE implementation, we grid-searched \(\beta \in \{1,5,10,20\}\) and found \(\beta=10\) optimal for the galaxy simulation dataset.  For Cars3D and dSprites, we adopted \(\beta=5\) following \cite{IDVAE}, given its proven balance of reconstruction fidelity and disentanglement. To be consistent with the settings of IDVAE, our experiments primarily focused on convolutional layers. A simpler MLP-based configuration is also provided in the Aux-VAE code repository for more general usage.

\begin{table}[htbp]
\centering
\caption{Architecture of the Aux-VAE for the Galaxy Simulation Dataset}
\label{tab:beta_vae_cnn_aux}
\begin{tabular}{|c|c|}
\hline
\textbf{Encoder} & \textbf{Decoder} \\
\hline
Input: $33 \times 33 \times 1$ & Input: $R^{10}$ \\
\hline
Conv2d(1, 32, 4, stride 2, padding 1), ReLU & ConvT2d(32, 128, 4, stride 2, padding 1), ReLU \\
\hline
Conv2d(32, 64, 4, stride 2, padding 1), ReLU & ConvT2d(128, 64, 4, stride 2, padding 1), ReLU \\
\hline
Conv2d(64, 128, 4, stride 2, padding 1), ReLU & ConvT2d(64, 32, 4, stride 2, padding 1), ReLU \\
\hline
Conv2d(128, 256, 4, stride 2, padding 1), ReLU & ConvT2d(32, 1, 5, stride 4, padding 0), Sigmoid \\
\hline
FC 256, FC $2 \times 10$ & \\
\hline
\end{tabular}
\end{table}

\begin{table}[h]
  \centering
  \begin{tabular}{l c c c}
    \hline
    \textbf{Dataset} & \(\beta\) & \(\lambda_1\) & \(\lambda_2\) \\
    \hline
    Galaxy simulation dataset & 5  & 1 & 0.1 \\
    Cars3D                    & 5  & 2 & 1   \\
    DSprites                  & 10 & 2 & 2   \\
    \hline
  \end{tabular}
  \caption{Hyperparameter values used for Aux-VAE on each evaluated dataset.}
  \label{tab:auxvae-hyperparams}
\end{table}
\clearpage


\end{document}